\documentclass{article}

\usepackage{microtype}
\usepackage{graphicx}
\usepackage{booktabs} % for professional tables
\usepackage{makecell}

\usepackage{hyperref}

\usepackage[accepted]{icml2023}

\usepackage{amsmath}
\usepackage{amssymb}
\usepackage{mathtools}
\usepackage{amsthm}
\usepackage[FIGTOPCAP]{subfigure}

\usepackage[capitalize,noabbrev]{cleveref}

\theoremstyle{plain}

\theoremstyle{definition}

\theoremstyle{remark}

\usepackage[textsize=tiny]{todonotes}

\icmltitlerunning{Mimetic Initialization of Self-Attention Layers}

\begin{document}

\twocolumn[
\icmltitle{Mimetic Initialization of Self-Attention Layers}



\icmlsetsymbol{equal}{*}

\begin{icmlauthorlist}
\icmlauthor{Asher Trockman}{xxx}
\icmlauthor{J. Zico Kolter}{xxx,yyy}
\end{icmlauthorlist}

\icmlaffiliation{xxx}{Carnegie Mellon University}
\icmlaffiliation{yyy}{Bosch Center for AI}

\icmlcorrespondingauthor{Asher Trockman}{ashert@cs.cmu.edu}

\icmlkeywords{Machine Learning, ICML}

\vskip 0.3in
]



\printAffiliationsAndNotice{}  % leave blank if no need to mention equal contribution

\begin{abstract}
\looseness=-1
It is notoriously difficult to train Transformers on small datasets;
typically, large pre-trained models are instead used as the starting point.
We explore the weights of such pre-trained Transformers (particularly for vision)
to attempt to find reasons for this discrepancy.
Surprisingly, we find that simply initializing the weights of self-attention layers 
so that they ``look'' more like their pre-trained counterparts
allows us to train vanilla Transformers faster and to higher final accuracies, 
particularly on vision tasks such as CIFAR-10 and ImageNet classification,
where we see gains in accuracy of over 5\% and 4\%, respectively.
Our initialization scheme is closed form, learning-free, and very simple:
we set the product of the query and key weights to be approximately
the identity, and the product of the value and projection weights
to approximately the negative identity.
As this mimics the patterns we saw in pre-trained Transformers,
we call the technique \emph{mimetic initialization}.

\end{abstract}

\section{Introduction}
\label{intro}
Despite their excellent performance in the regime of large-scale pretraining,
Transformers are notoriously hard to train on small-scale datasets~\cite{vit}.
In this setting, convolutional networks such as the ResNet tend to massively outperform 
Vision Transformers, with the gap only being closed by the addition of techniques such
as self-supervised pretraining, auxiliary losses, convolution-inspired tokenizers,
or the addition of other architectural components that promote convolution-like inductive biases.
Similar effects are seen in language modeling, where classic models such as LSTMs outperform
vanilla Transformers without extreme regularization and long-duration training.

\looseness=-1
In this work, we take a step towards bridging this gap via a novel initialization technique
for Transformers. We focus primarily on Vision Transformers (ViTs), though we also investigate our technique
in the context of language modeling.
We note that in pretrained ViTs, 
the weights of self-attention layers are often quite correlated,
in that $W_Q W_K^T \propto I + \epsilon$ and $W_V W_{proj} \propto \epsilon - I$.
Our proposal is merely to initialize the self-attention weights to mimick this observation,
with the added caveat of requiring standard sinusoidal position embeddings.
While we propose only one technique here, we believe that this concept
is worthy of future research, as it may enhance the understanding of 
the inner-workings of deep models and lead to cheaper training and better optima.
We propose to call this type of technique \emph{mimetic initialization}, as we initialize 
by \emph{mimicking} the structures and patterns observed in the weights of \emph{pretrained} models.
Importantly, the sort of \emph{mimetic initialization} we propose seeks to mimic \emph{solely} through
hand-crafted, interpretable formulas: it involves absolutely no pretraining and is practically compute-free;
\emph{i.e.,} there is no learning procedure involved.

Fundamentally, we seek to investigate the question proposed by \citet{lego}:
might some of the benefits of pretraining actually just be a result of it serving as a good initialization?
Our approach is to attempt to find good initializations that do not involve pretraining to begin to explore this question.

Our initialization shows strong advantages for ViTs, allowing gains of up to 5\% when training
on small datasets like CIFAR-10, and up to 4\% for larger datasets, \emph{i.e.,} ImageNet-1k within a standard
ResNet-style training pipeline.
We also see smaller performance gains on language modeling tasks such as WikiText-103.

\begin{figure*}
\centering
	\subfigure[$W_Q W_K^T$ often has a noticeable positive diagonal. $\rightarrow$ Layers 1-12, $\downarrow$ Attention Heads 1-3 ]{\includegraphics[width=\textwidth,trim=0 0 0 0,clip]{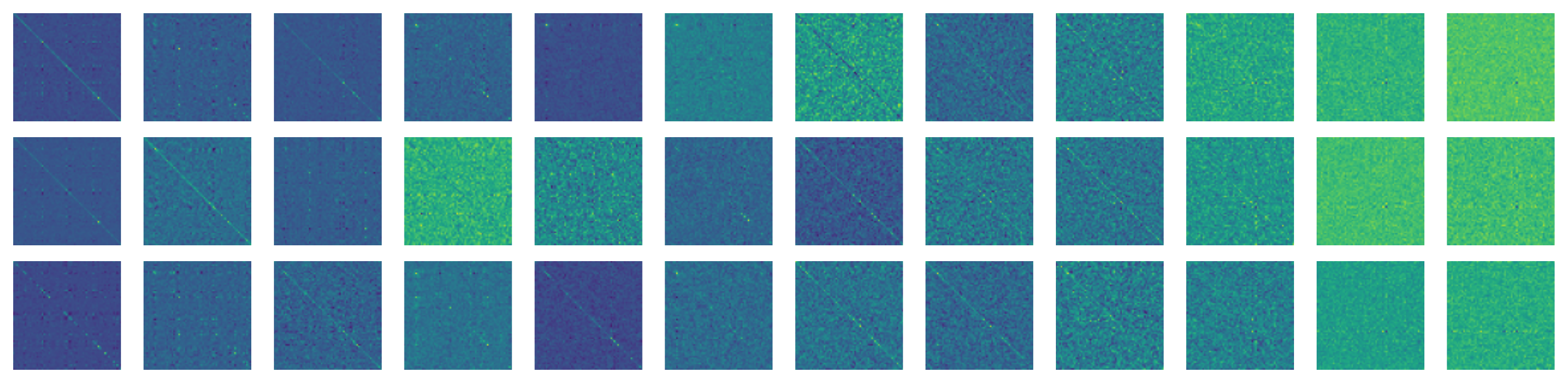}}
    \subfigure[$W_V W_{proj}$ often has a prominent negative diagonal. Here, we sum over heads.]{\includegraphics[width=\textwidth,trim=0 0 0 0,clip]{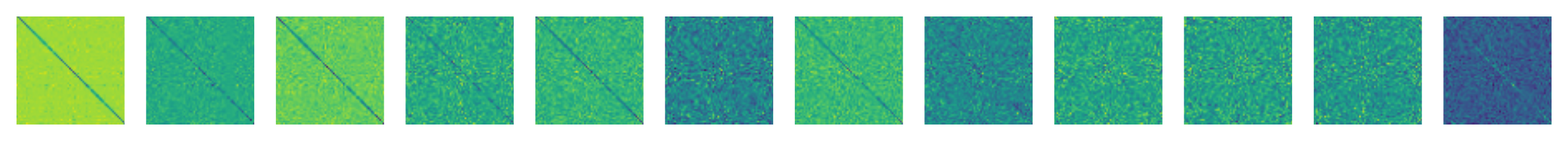}}
\vspace*{-0.5em}
\caption{Self-attention weights of an ImageNet-pretrained ViT-Tiny. Pictured are 3 heads for each of the 12 layers. Clipped to 64x64.}
\label{ref:vit-observations}
\end{figure*}

\section{Related Work}

It is conventional wisdom that CNNs have a stronger inductive bias
than ViTs. In practice, this means that CNNs perform particularly well on small datasets,
while ViTs only surpass their performance when pretrained on very large (\emph{e.g.,} ImageNet-21k- or JFT-300B-scale)
datasets.
To remedy this situation, numerous works have proposed to integrate convolutions explicitly into ViTs:
\citet{coatnet} introduces CoAtNet, which directly integrates depthwise convolution and self-attention.
\citet{cvt} introduces CvT, a Transformer modification involving convolutional tokenization and projections.
\citet{ceit} proposes the Convolution-enhanced Image Transformer (CeiT), which makes various modifications
to bring about CNN-like inductive bias.
These techniques are uniformly effective: ViT/CNN hybrids tend to achieve higher accuracies with less data than their
vanilla ViT counterparts.
In contrast to these works, we seek to make ViTs more trainable \emph{without} the use of convolutions,
guided by the observation that pretrained ViTs eventually become effective without them given sufficient training time.

There are relatively few works on initializing Transformers;
these works tend to be theoretical, focusing on eliminating normalization or skip connections.
\citet{tfixup} investigates training Transformers without learning rate warmup and normalization,
and proposed a rescaling of weights that allows these to be removed.
\citet{dks} extends work on Deep Kernel Shaping to train Transformers without normalization
and skip connections.
Rather than initializing $W_Q, W_K$ in a particularly structured or principled way, they
ensure the product is zero and instead add a controllable bias inside the softmax of the self-attention layers.
Similarly, \citet{zero} proposes to set the query and key weights to zero and the identity, respectively;
however, the product of these weights remains zero.

In contrast, we attempt to better-initialize standard \emph{vanilla} Transformers, which 
use skip connections and normalization.
Moreover, we do so by controlling the behavior of the query and key weights themselves,
aiming to replicate the behavior of pretrained models without any training.

\citet{layerscale} proposes LayerScale, which multiplies the skip connections by a learnable diagonal matrix;
though this is an actual architectural change and not an initialization, we will discuss the potential (albeit weak) connection
to our initialization in Sec.~\ref{sec:why-works}.
\citet{cordonnier} and \citet{convit} propose a scheme to initialize self-attention to implement convolution;
however, this requires the use of relative positional embeddings and the (gated) self-attention layers proposed must have a particular number of heads
to match the kernel size.
In contrast, our scheme makes no architectural changes to the Transformer and still achieves
comparable performance.
Importantly, we do not seek to make self-attention emulate convolution explicitly, 
but rather emulate the behavior of self-attention \emph{itself}
after large-scale pretraining.

An inspiration for our work, \citet{lego} proposed a so-called ``mimicking initialization'' as an alternative to large-scale pretraining
for language models. However, this technique actually \emph{trains} self-attention layers to mimick the behavior of
a handcrafted, convolution-like target similar to attention maps seen in trained models; in contrast, we attempt to bring about 
desirable behavior of self-attention entirely by hand, without any form of training.
In that sense, our method is vaguely similar in spirit to \cite{convcov}, who propose a learning-free, structured multivariate
initialization for convolutional filters.

Many works have modified Vision Transformers to more effectively train on small-scale datasets.
\citet{gani2022train} proposes to learn the weight initialization in a self-supervised fashion, noting that ViTs are highly sensitive to initialization. This achieves good results on CIFAR-10 and other small-scale datasets.
\citet{cao2022training} proposes another self-supervised technique for from-scratch training.
\citet{hassani2021escaping} proposes a Compact Convolutional Transformer that can perform well on small datasets, which involves the use of a convolutional tokenizer.
\citet{lee2021vision} improves performance on small-scale datasets by introducing Shifted Patch Tokenization and Locality Self-Attention.
\citet{liu2021efficient} proposes a ``dense relative localization'' auxiliary task which  
improves the performance of transformers on small-scale datasets.
In contrast to these works, which introduce auxiliary tasks or novel components, we use standard ResNet-style training and still achieve
good results on small datasets with \emph{completely vanilla Transformers}.

\section{Observations}
\label{sec:observations}

\looseness=-1
\paragraph{Preliminaries}
We denote the query and key weight matrices for a single head of self-attention by
$W_Q, W_K \in \mathbb{R}^{d \times k}$,
where $d$ is the dimension (or width) of the Transformer
and $k = d / \# \text{heads}$ is the head dimension.
We consider the value and projection matrices to be full-rank:
$W_V, W_{proj} \in \mathbb{R}^{d \times d}$. For inputs $X \in \mathbb{R}^{n \times d}$
with additive positional embeddings $P \in \mathbb{R}^{n \times d}$,
we denote the ``attention map'' as follows:
$$
\mathsf{Softmax}\left(\tfrac{1}{\sqrt{k}} X W_Q W_K^T X^T \right).
$$

\looseness=-1
Our initialization is based on \emph{mimicking} the patterns we observed
in pre-trained vision transformers.
In Fig.~\ref{ref:vit-observations}, we visualize said patterns
for a ViT-Tiny, pretrained on ImageNet.
The diagonal of the product of $W_Q$ and $W_K^T$ is noticeably positive in many cases.
Similarly, and somewhat surprisingly, the product of $W_V$ and $W_{proj}$ tends to
have a noticeably negative diagonal.
This similarly holds for ViTs of different sizes.
This suggests that, in rough approximation,
$W_Q$ and $W_K$ may be the ``same'' low-rank random normal matrix,
as such matrices are approximately semi-orthogonal.
This is based on the fact that an appropriately-scaled random normal matrix
is approximately orthogonal.
That is, if $Z \in \mathbb{R}^{d \times k}$ and $Z \sim \mathcal{N}(0, I/k)$,
then $ZZ^T \approx I$.
On language models (see Fig.~\ref{ref:tmr-observations}),
we see a similar, albeit not quite so clear pattern.
In contrast, the products $W_Q$ and $W_K^T$ are often negative instead of positive,
and vice versa for $W_V$ and $W_{proj}$.

\begin{figure}
\centering
	\subfigure[]{\includegraphics[width=0.19\columnwidth,trim=0 0 0 0,clip]{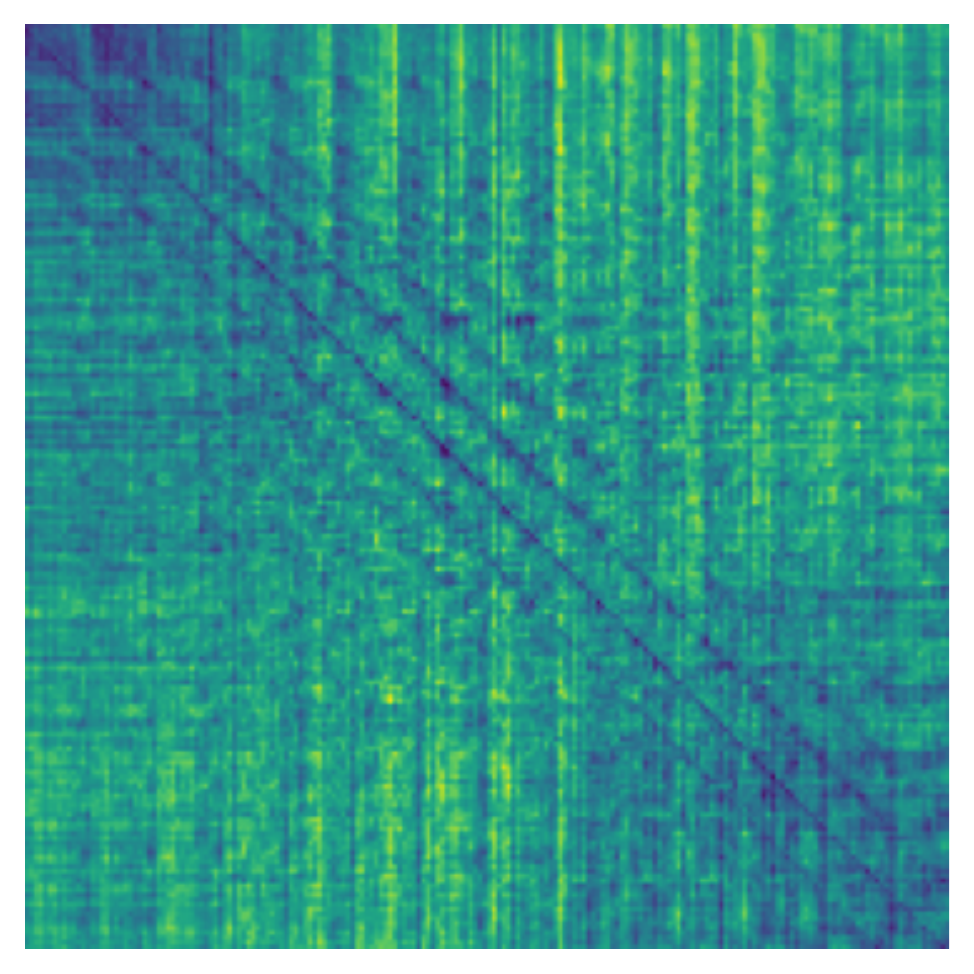}}
	\subfigure[]{\includegraphics[width=0.19\columnwidth,trim=0 0 0 0,clip]{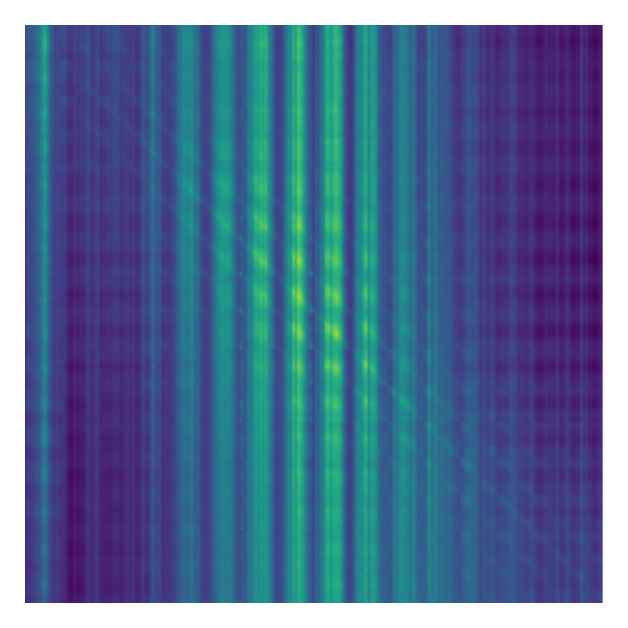}}
	\subfigure[]{\includegraphics[width=0.19\columnwidth,trim=0 0 0 0,clip]{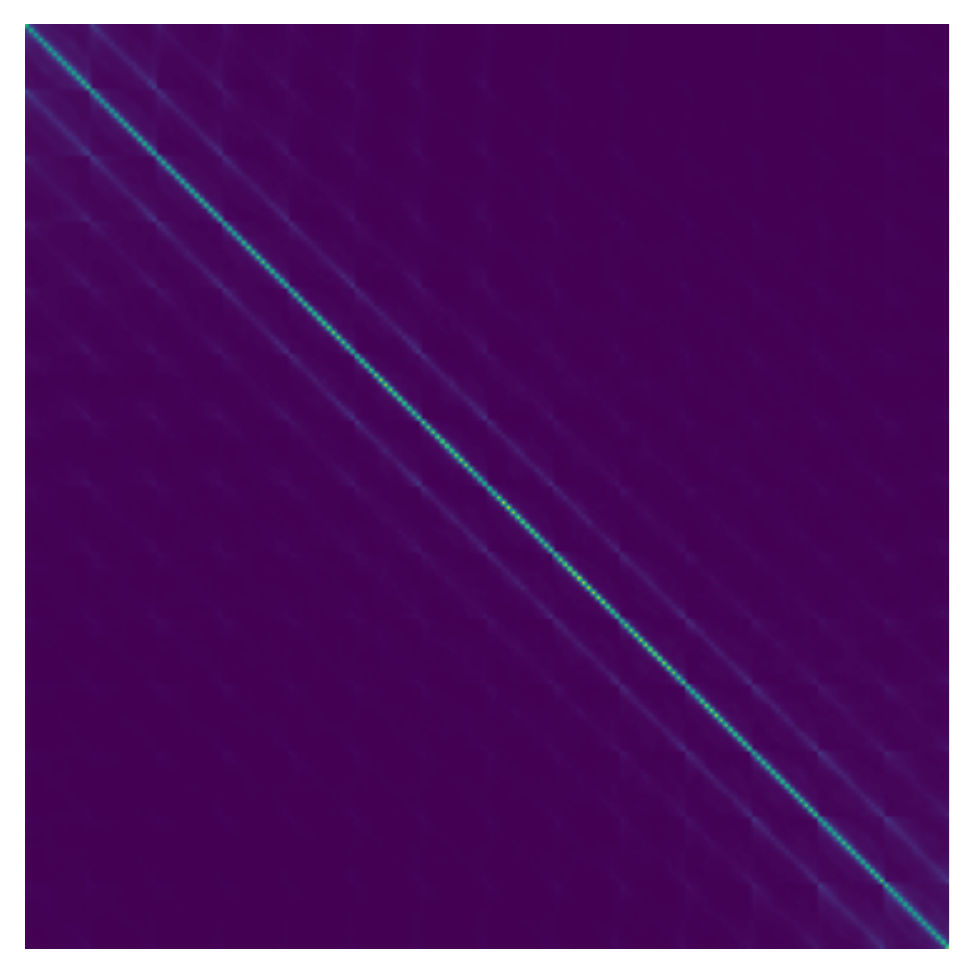}}
	\subfigure[]{\includegraphics[width=0.19\columnwidth,trim=0 0 0 0,clip]{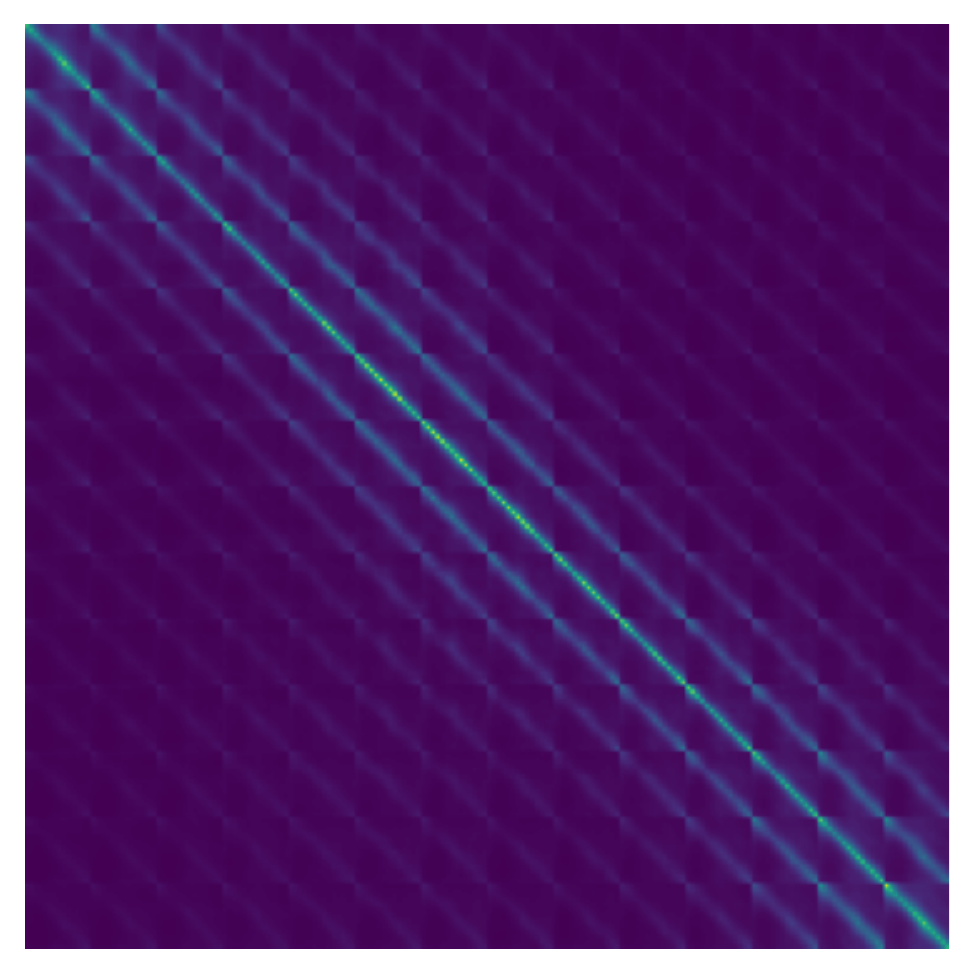}}
	\subfigure[]{\includegraphics[width=0.19\columnwidth,trim=0 0 0 0,clip]{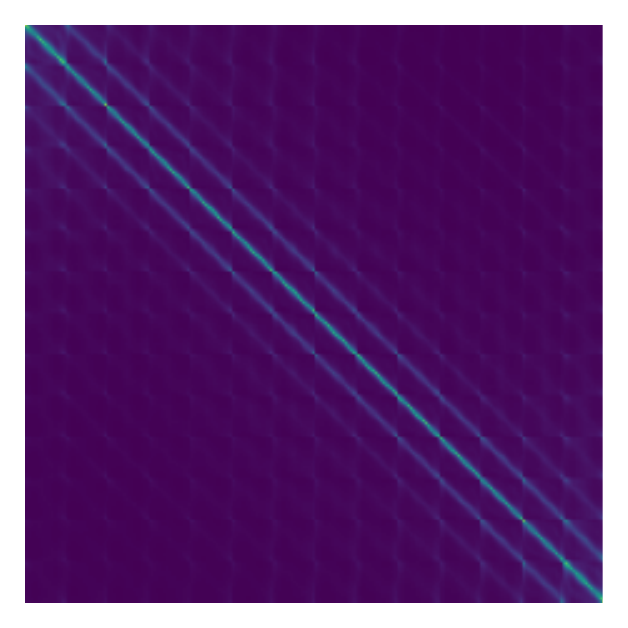}}

	\subfigure{\includegraphics[width=0.19\columnwidth,trim=0 0 0 0,clip]{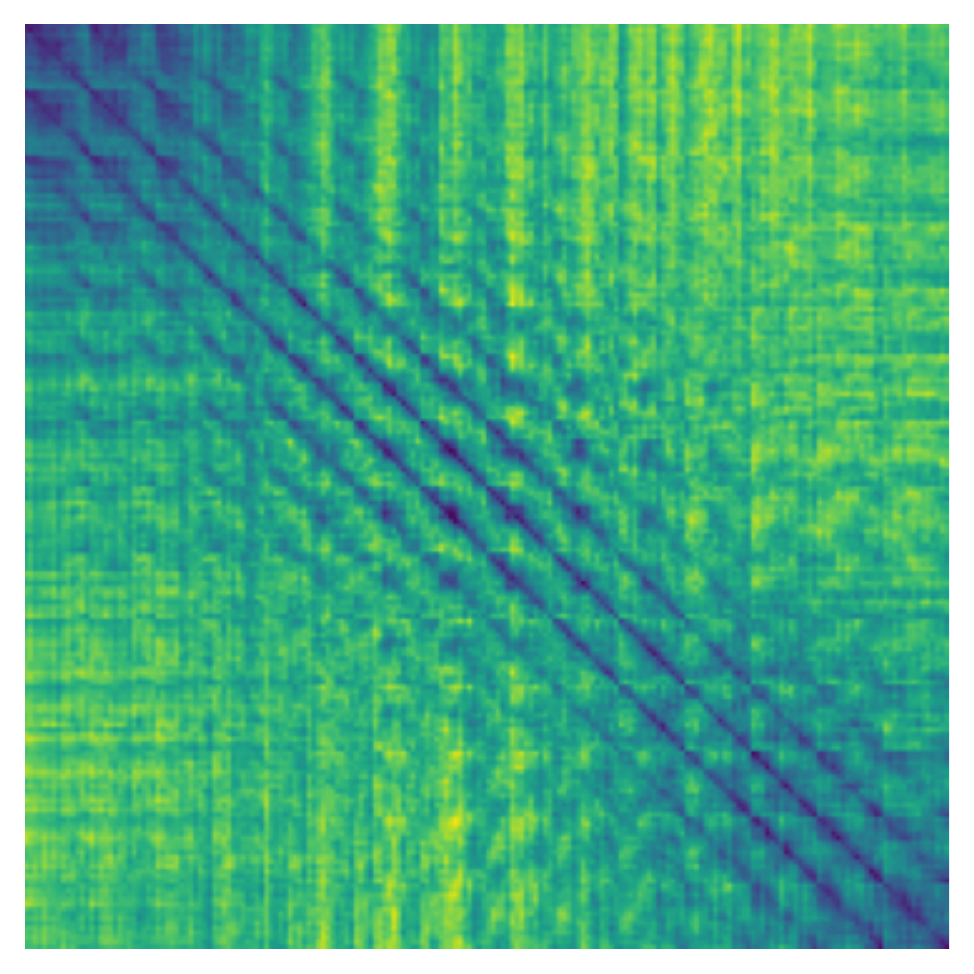}}
	\subfigure{\includegraphics[width=0.19\columnwidth,trim=0 0 0 0,clip]{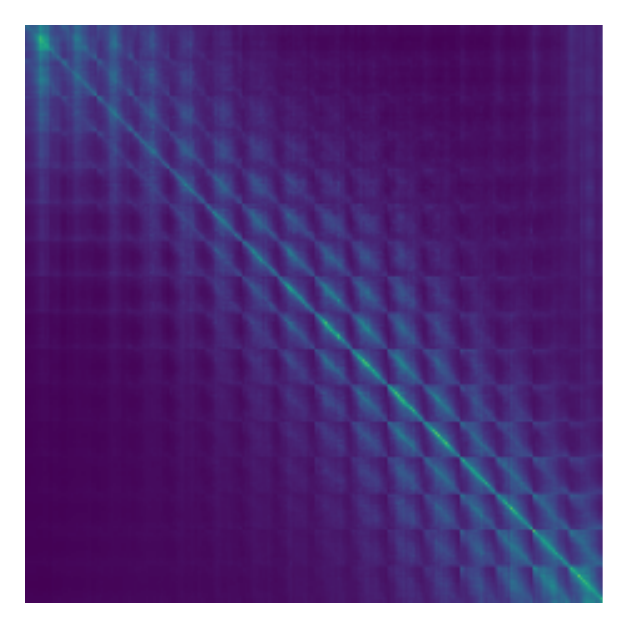}}
	\subfigure{\includegraphics[width=0.19\columnwidth,trim=0 0 0 0,clip]{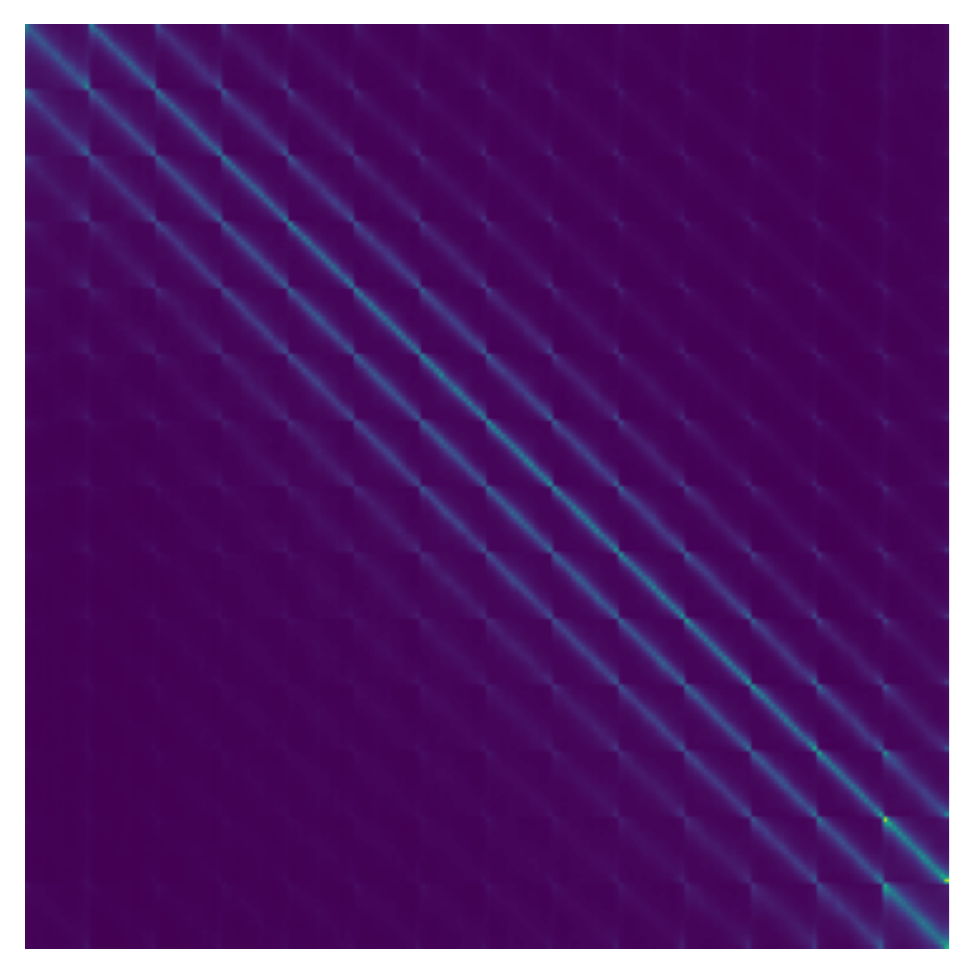}}
	\subfigure{\includegraphics[width=0.19\columnwidth,trim=0 0 0 0,clip]{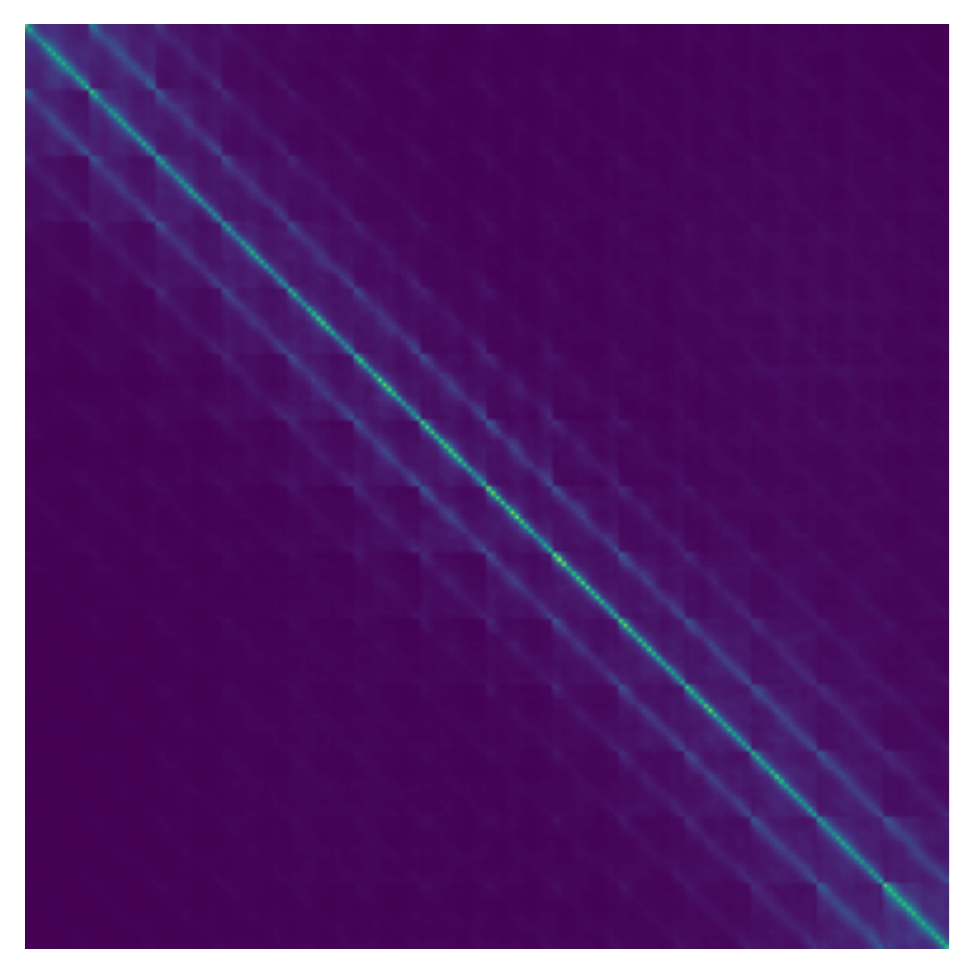}}
	\subfigure{\includegraphics[width=0.19\columnwidth,trim=0 0 0 0,clip]{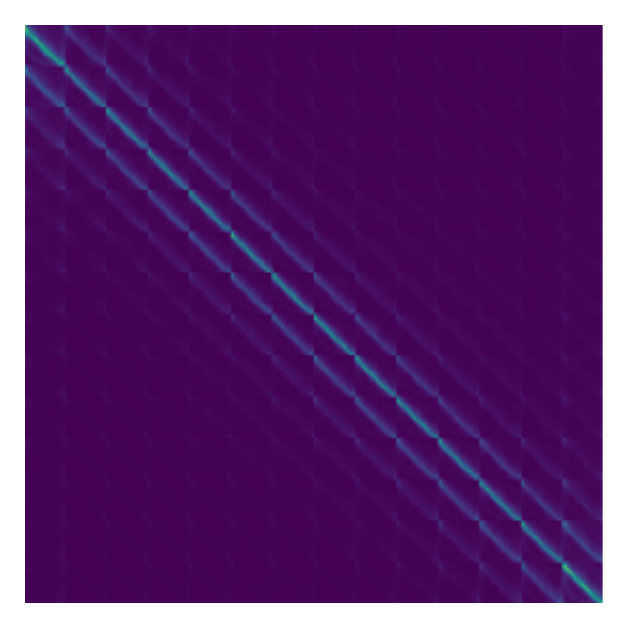}}

	\subfigure{\includegraphics[width=0.19\columnwidth,trim=0 0 0 0,clip]{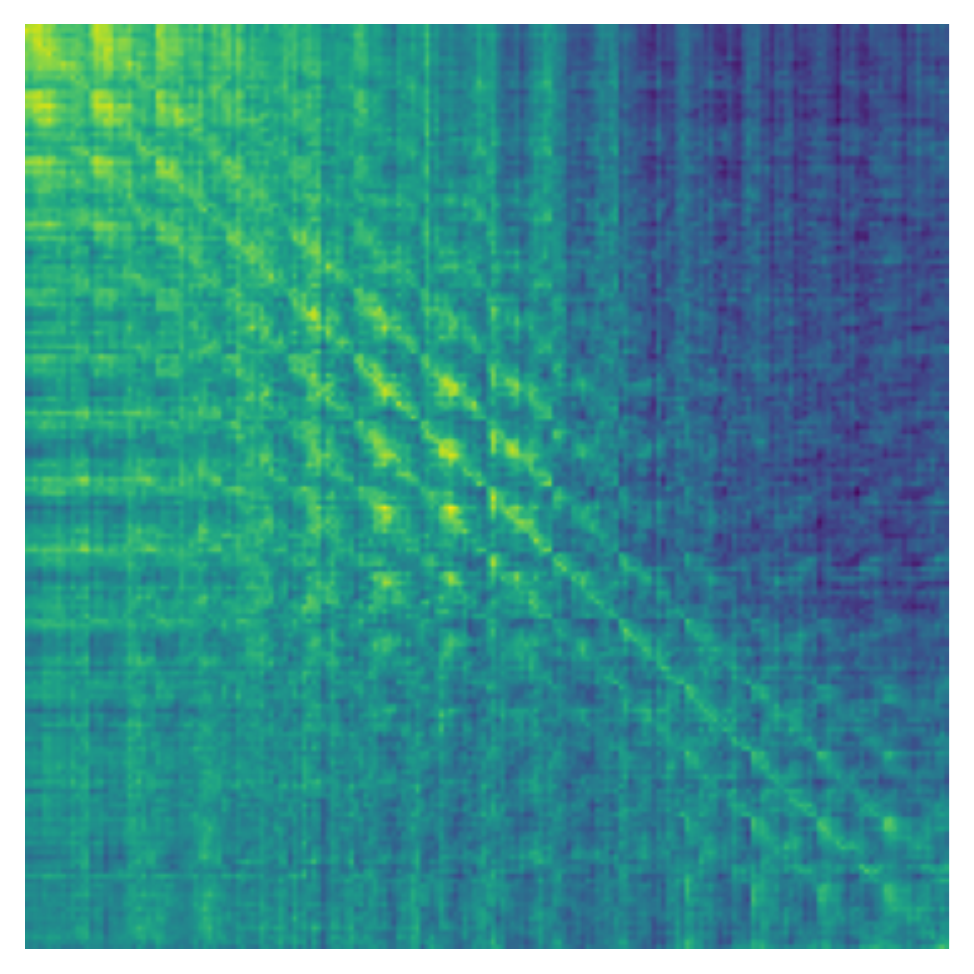}}
	\subfigure{\includegraphics[width=0.19\columnwidth,trim=0 0 0 0,clip]{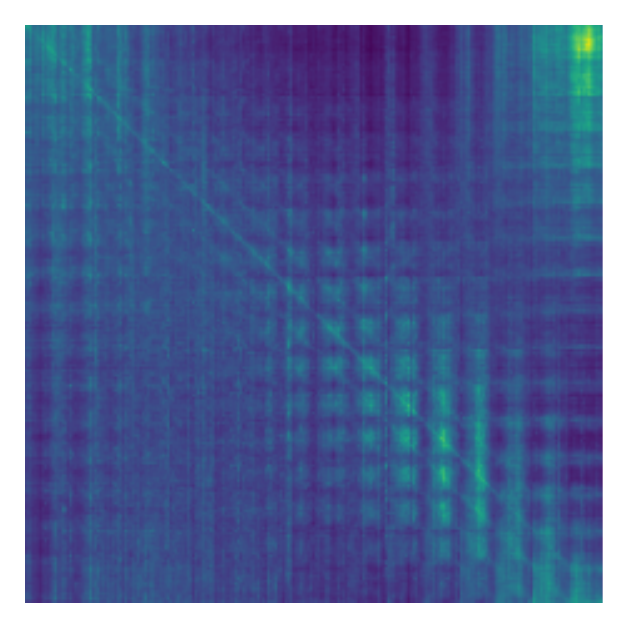}}
	\subfigure{\includegraphics[width=0.19\columnwidth,trim=0 0 0 0,clip]{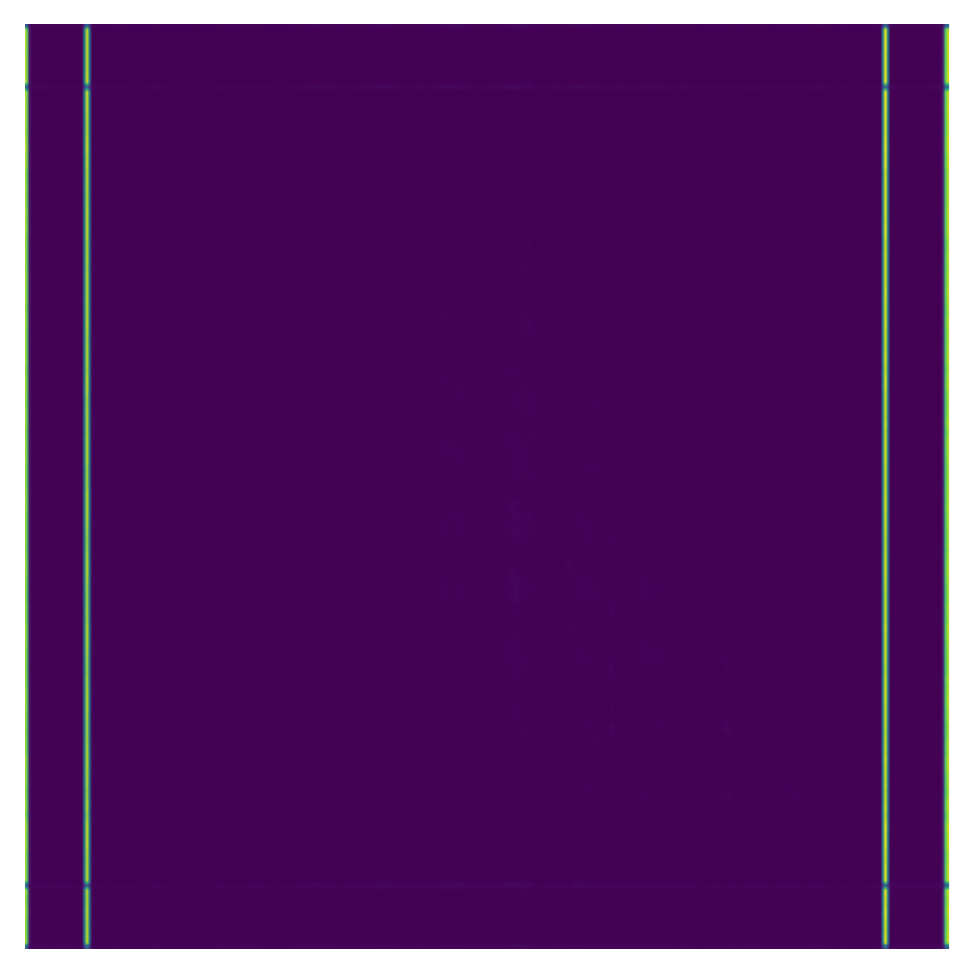}}
	\subfigure{\includegraphics[width=0.19\columnwidth,trim=0 0 0 0,clip]{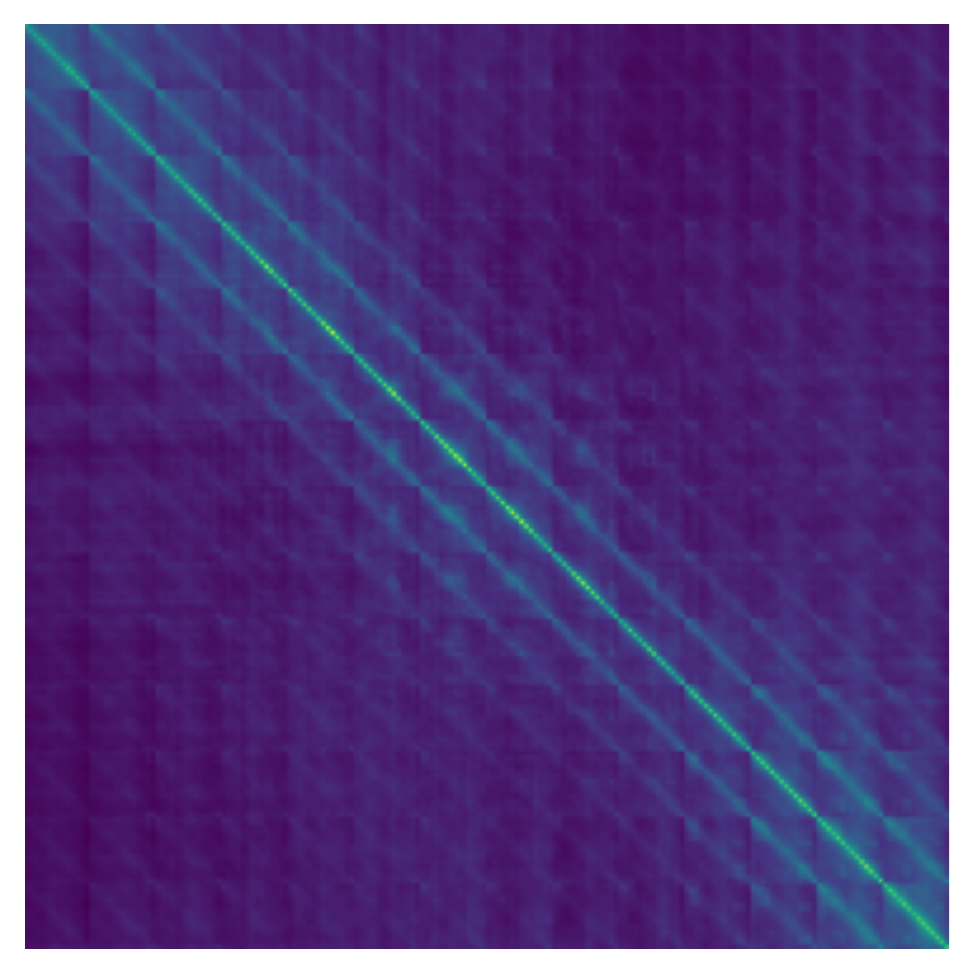}}
	\subfigure{\includegraphics[width=0.19\columnwidth,trim=0 0 0 0,clip]{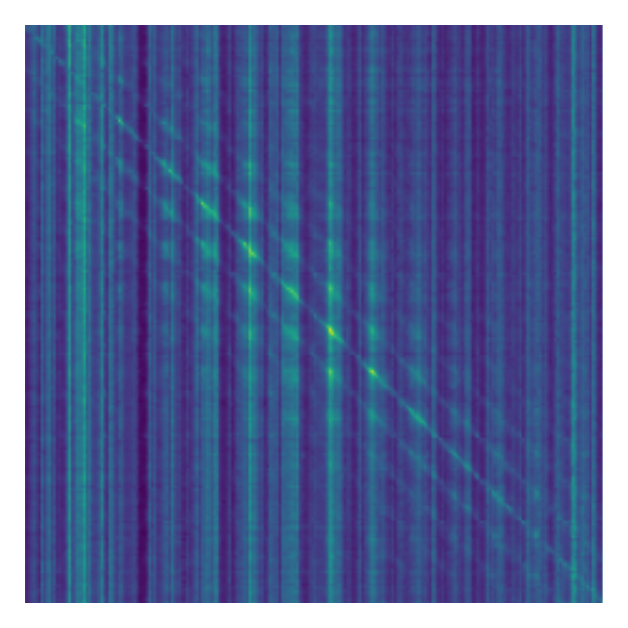}}
    %\vspace*{-0.5em}
	\caption{Attention maps computed from one CIFAR-10 batch for ViT-Tiny (a) untrained (b) CIFAR-10 trained (c) ImageNet pretrained (d) using our init (e) our init and then CIFAR-10 trained. \newline Rows: $\downarrow$ Layers \#1, 4, 11}
	\label{ref:attn-maps}
\end{figure}

In Figure~\ref{ref:attn-maps}, we show the attention maps in a ViT-Tiny for a variety
of training settings,
averaged over the three heads and over a batch of CIFAR-10 inputs.
Note the difference between the untrained model (a) and the untrained one using our initialization (d).
Further, there is some degree of similarity between the ImageNet-pretrained model (c)
and our untrained one (d). After training our initialized ViT on CIFAR-10, the early layers are
similar to those of the ImageNet-pretrained ViT while the later layers are more like
those of the only-CIFAR-trained ViT (b).
The last layers of the ImageNet-pretrained ViT implement a kind of broadcasting operation
which we do not attempt to mimick.

\section{Method}

We note that there are two relatively simple choices
for modeling $W_Q W_K^T$ and $W_V W_{proj}$.
The simplest technique is to merely set the two matrices
in the product to the same random normal matrix, \emph{i.e.},
$W_Q = W_K = N(0, I/k)$, which is scaled by the Transformer
head dimension $k$ so that the average magnitude of the diagonal is $\approx 1$.
In the case of the value/projection matrices, whose diagonal we
want to be negative, this would be $$Z \coloneqq N(0, I/d), W_V = Z, W_{proj} = -Z.$$
However, no matter how we scale the random normal matrix, the ratio between
the magnitude of the on-diagonal and the off-diagonal noise 
remains the same.

\looseness=-1
To gain more flexibility in the prominence of the diagonal, we instead propose
to use a slightly more involved technique. Here, we explicitly model
the products as follows:
\begin{align}
	W_Q W_K^T \approx \alpha_1 Z_1 + \beta_1 I \\
	W_V W_{proj}^T \approx \alpha_2 Z_2 - \beta_2 I
\end{align}
where $Z_i \sim \mathcal{N}(0, \tfrac{1}{d}I)$ and
$\alpha_i, \beta_i \in [0, 1]$.
That is, we explicitly control the tradeoff between the noise $Z_i$
and the diagonal $I$ by choosing the parameters $\alpha_i, \beta_i$.
In order to recover the factors $W_V, W_{proj}$, we use the singular value decomposition: 
\begin{align}
&\alpha_1 Z_1 + \beta_1 I = U_1 \Sigma_1 V_1^T \\
&W_V \coloneqq U_1 \Sigma_1, W_{proj} \coloneqq V_1 \Sigma_1^{1/2},
\end{align}
and for the low-rank factors $W_Q, W_K^T$, the reduced SVD:
\begin{align}
&\alpha_2 Z_2 + \beta_2 I = U_2 \Sigma_2 V_2^T \\
&W_Q \coloneqq U_2[:,:k] {\Sigma_2[:k,:k]}^{1/2} \\
&W_K \coloneqq V_2[:, :k] \Sigma_2[:k, :k]^{1/2}.
\end{align}
Note that we resample $Z_2$ for each head.

\begin{figure}
	\centering
\includegraphics[width=\columnwidth,trim=0 0 0 0,clip]{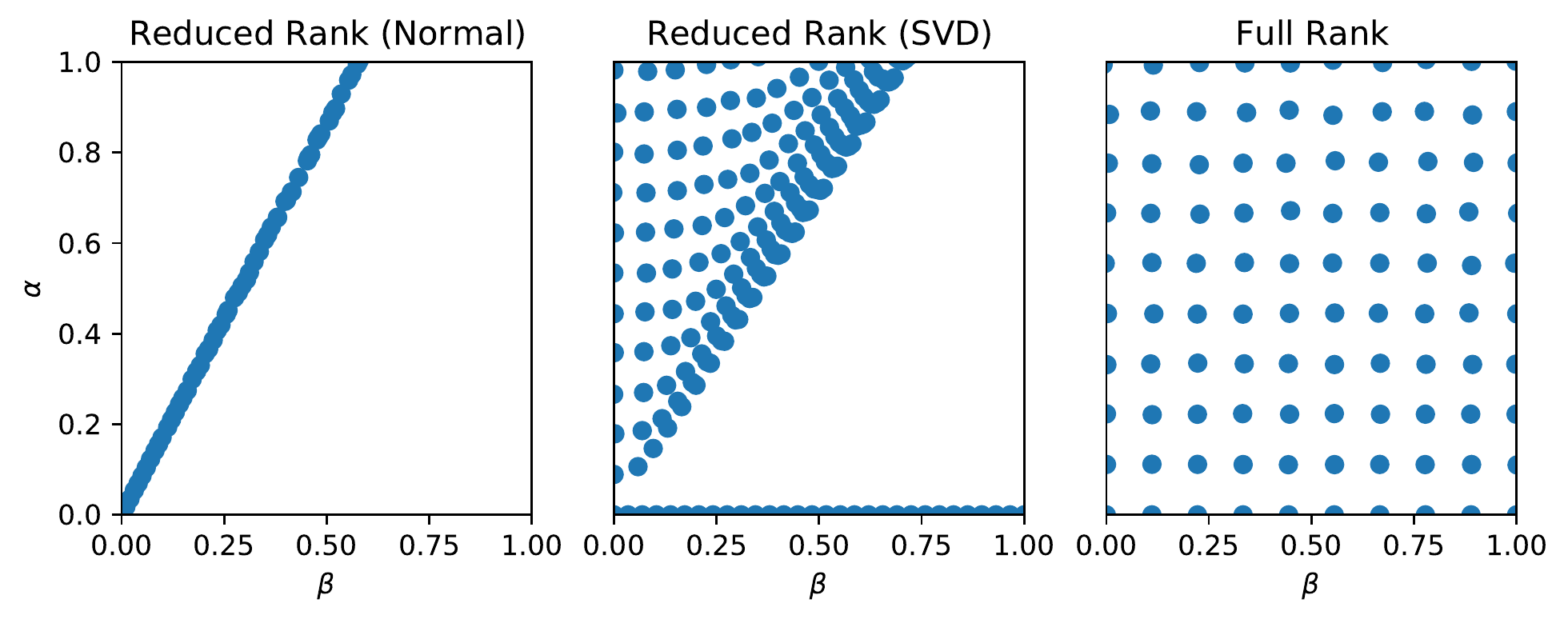}
	\vspace{-2em}
	\caption{Possible $\alpha, \beta$ for different weight constructions.}
	\label{fig:alphabeta}
\end{figure}

In Fig.~\ref{fig:alphabeta}, we show the different $\alpha, \beta$ that can be achieved
through the two methods proposed above. Using equal random normal matrices,
there is a linear relationship between $\alpha$ and $\beta$, for both low-rank
and full-rank matrices. Using the SVD technique, we achieve a wider variety
of selections even in the low-rank case.
Consequently, we use this in all experiments.

\paragraph{Attention map structure}
In practice, our initialization results in attention maps
that with a strong diagonal component which reflect the structure 
of the position embeddings, which we denote by $P \in \mathbb{R}^{n \times d}$. 
We show this visually in Fig.~\ref{ref:attn-maps}, though it is also possible
to (roughly) compute their expected value.

Assuming that $X \in \mathbb{R}^{n\times d}$ and $X \sim \mathcal{N}(0, I)$ (which is a reasonable
assumption due to the use of LayerNorm),
and assuming $W_Q, W_K$ are full-rank and $W_Q W_K^T = \alpha Z + \beta I$ due to our initilization,
we can show $\mathbb{E}[(X + P) (\alpha Z + \beta I) (X + P)^T] = \beta d I + \beta PP^T$,
as the only products with non-zero mean are $XX^T \approx I$ (on the diagonal)
and $PP^T$.
Thus, roughly speaking, our initialization results in expected attention maps
of the form
\begin{equation}
\label{eq:near-eye}
    \mathsf{Softmax} \left(\tfrac{1}{\sqrt{k}} ( \beta_1 d I + \beta_1 PP^T) \right).
\end{equation}
That is,
our initialization may bias attention maps towards
mixing nearby tokens according to the structure of $PP^T$,
which can be seen in Fig.~\ref{ref:attn-maps}.

\begin{table}[t]
\caption{100 epoch CIFAR-10 classification (ViT-Tiny).}
\label{table:cifar}
\vskip 0.15in
\begin{center}
\begin{small}
\begin{tabular}{lccccr}
\toprule
    \thead{Width} & \thead{Depth} & \thead{Heads} & \thead{Acc. \\ (Base)} & \thead{Acc. \\ (Init)} & \thead{$\Delta$ Acc.} \\
\midrule
96  & 6  & 3  & 84.75& 87.90 & 3.15  \\
96  & 12 & 3 & 84.75 & 88.84 & 4.09   \\
192 & 6  & 3 & 85.85 & 89.68  & 4.63   \\
192 & 12 & 1  & 85.25& 89.88 &  4.63 \\
192 & 12 & 3  & 86.07& 90.78 & 4.71 \\ 
192 & 12 & 6  & 86.74& 91.38 & 4.64  \\
192 & 24 & 3  & 86.36& 91.85 & 5.49 \\
384 & 12 & 3 & 86.26 & 91.56 & 5.30 \\
384 & 12 & 6  & 84.40& 92.17 & 7.77 \\
384 & 12 & 12 & 86.39& 92.30 & 5.91 \\
\bottomrule
\end{tabular}
\end{small}
\end{center}
\vskip -0.1in
\end{table}

\section{Experiments}

\subsection{CIFAR-10}

\looseness=-1
Training vanilla ViTs from scratch on CIFAR-10 is notoriously difficult,
requiring semi-supervised pretraining techniques, additional inductive bias,
or heavy data augmentation with long training times~\cite{liu2021efficient, lee2021vision, gani2022train, hassani2021escaping}.
In this section, we demonstrate the substantial benefits of using our initialization for vanilla
ViTs on from-scratch CIFAR-10 training.

\paragraph{Setup} We train all ViTs using a simple pipeline: we use RandAugment and Cutout for augmentation,
a batch size of 512, AdamW with $3\times10^{-3}$ learning rate, 0.01 weight decay, and 100 epochs.
We use a vanilla ViT with embedding dimension 192, depth 12, patch size 2, and input size 32 unless
otherwise noted (ViT-Tiny). We use a class token and sinusoidal position embeddings.
We use $\alpha_1 = \beta_1 = 0.7$
and $\alpha_2 = \beta_2 = 0.4$ for all experiments.

\paragraph{Basic results}
In Table~\ref{table:cifar}, we show our main results for CIFAR-10.
Across a variety of ViT design parameters, our initialization results in 
substantial accuracy gains between 2.5-6\%.
While the benefit of our initialization is quite significant in all cases,
we note that it seems to have the most benefit for larger models.
For example, we see an improvement of over 6\% for a ViT with dimension (width)
384, depth 12, and 6 heads (a ViT-Small), while we see a smaller 4.8\% gain
for a model with dimension 192 and 3 heads, and a 4.1\% gain for dimension 96.

\paragraph{Ablations}
In Table~\ref{table:ablate}, we show some ablations of our initialization technique.
If we use the default normal initialization for $W_Q, W_K$, we see a substantial loss of accuracy of nearly 2\%;
similarly, if we use default initialization for $W_V, W_{proj}$, we see an even greater hit to accuracy of around 3.5\%.
Using neither (just sinusoidal position embeddings), we lose almost 4\% accuracy.
Further, setting the diagonal of $W_V, W_{proj}$ to be negative rather than positive is in fact quite important,
accounting for around 1.5\% accuracy.
These results suggest that all of the components of our initialization work together, and all are very important.
We note that in Fig.~\ref{ref:vit-observations} the prominence of the diagonal tends to fade with depth;
we saw no improvement from mimicking this.

\paragraph{GPSA comparison}
GPSA (Gated Positional Self-Attention) was proposed for use in the ConViT model
by \citet{cordonnier}. This self-attention variation has two attention maps,
one of which is initialized with ``soft'' convolutional inductive biases
to emulate convolution. The effect of each attention map is
determined by a learnable gating parameter.

While our goal was to improve Transformers without architectural modifications,
this technique is the most similar to our own. (Though it requires, \emph{e.g.,}
a particular number of heads and a new, custom layer.)
We replaced all self-attention layers with GPSA layers.
With 4 heads (approximately 2x2 convolution),
accuracy comes fairly close to our own by around 0.6\%.
Interestingly, adding our $W_V W_{proj}$ initialization to GPSA further 
narrows the gap by around 0.4\%.
This shows that our technique may even be useful for self-attention variants.
More importantly, it shows that our technique is competitive even
with those requiring more extensive architectural changes or 
explicitly-constructed convolutional biases.

\begin{table}
\caption{Ablations on CIFAR-10, ViT-Ti}
\label{table:ablate}
\vskip 0.15in
\begin{center}
\begin{small}
\begin{tabular}{lc}
\toprule
\thead{Ablation} & \thead{Acc.}\\
\midrule
Our initialization & 91.38 \\
\midrule
Random pos. embeddings & 88.70 \\
No init (only sinusoidal pos. embeddings) & 87.39 \\
Init only $W_Q, W_K$ & 89.17 \\
Init only $W_V, W_{proj}$ & 87.23 \\
$W_V W_{proj} \propto -cI$ $\Longrightarrow$ $W_V W_{proj} \propto +cI$ & 89.65 \\
\midrule
GPSA (8 heads) & 90.03 \\
GPSA (4 heads) & 90.83 \\
\; + $W_V W_{proj} \propto -cI$ & 91.21 \\
\midrule
\makecell[l]{Pretrained \\ $W_K, W_Q, W_V, W_{proj}$ \& pos. embed} & 91.15 \\
\bottomrule
\end{tabular}
\end{small}
\end{center}
\vskip -0.1in
\end{table}

\paragraph{Pretrained weights}
Our initialization technique only considers position embeddings and the
query, key, value, and projection weights.
Consequently, we consider transfering just these weights from an ImageNet-pretrained
ViT as a baseline initialization technique.
This achieves 91.15\% accuracy, which is marginally lower than our own initialization.
While this does not say anything about the initialization of the patch embedding and MLP layers,
this may provide some evidence that our self-attention initialization is close to optimal.

\paragraph{Position embeddings}

According to  Table~\ref{table:ablate}, the use of sinusoidal position embeddings instead
of randomly-initialized ones is crucial for our initialization.
Using random rather than sinusoidal position embeddings with our initialization
is disastrous, resulting in a decrease of 3\% in accuracy.
However, \emph{only} initializing the position embeddings is not helpful either;
ablating the rest of the init gives a similar performance decrease.
In other words, it is the \emph{interaction}
of our initialization with the position embeddings which is useful.
Consequently, with Eq.~\ref{eq:near-eye} in mind, we investigated the scale of the position embeddings,
which changes their importance relative to the inputs themselves.

\begin{figure}
	\includegraphics[width=0.25\textwidth,trim=0 0 0 0,clip]{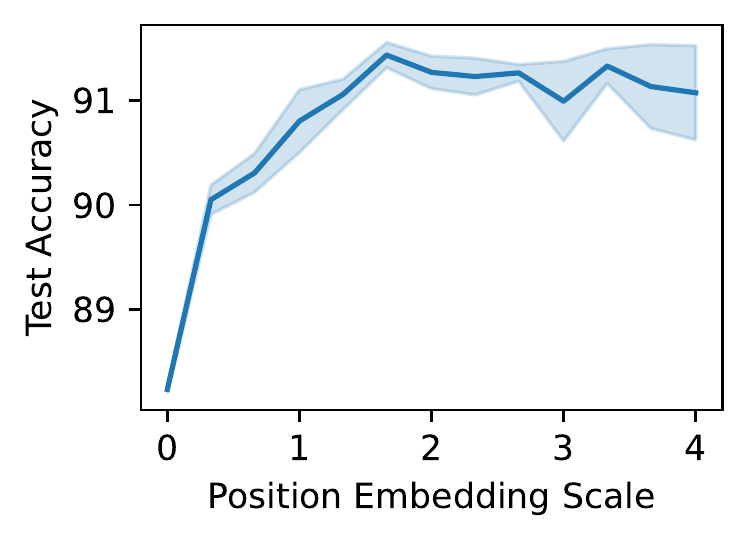}
	\centering
	\caption{Increasing the scale of the position embeddings improves CIFAR-10 performance (ViT-Tiny).}
	\label{fig:posemb}
\end{figure}

\paragraph{Position embedding scale} 
Adding a new hyperparameter, we multiplied the embeddings by a factor $\gamma$, and tried
several choices as shown in Fig.~\ref{fig:posemb}.
Increasing the scale from 1 to $\approx 2$ substantially improves performance,
by around 0.5\%.

\begin{table}[t]
\caption{ImageNet Results}
\label{sample-table}
\vskip 0.15in
\begin{center}
\begin{small}
\begin{tabular}{lcccccr}
\toprule
	\thead{Arch.} & \thead{Patch \\ Size}  & \thead{Batch \\ Size} & \thead{Input \\ Size} & \thead{Acc. \\ (Base)} & \thead{Acc. \\ (Init)} & \thead{$\Delta$\\ Acc.} \\
\midrule
\multicolumn{7}{c}{$\downarrow$ ResNet-style Training Pipeline (150 epochs) $\downarrow$ } \\
\midrule
Vit/Ti  & 16 & 640 & 224 & 70.28 & \textbf{73.08} & 2.8  \\
Vit/Ti  & 16 & 1024 & 224 & 67.80 & \textbf{71.92} & 4.1  \\
\midrule
\multicolumn{7}{c}{$\downarrow$ DeiT-style Training Pipeline (300 epochs)  $\downarrow$ } \\
\midrule
Vit/Ti  & 16 & 1024 & 224 & 72.08 & \textbf{72.65} & 0.57  \\
Vit/S  & 16 & 1024 & 224 & 79.83 & \textbf{80.36} & 0.53  \\
\bottomrule
\end{tabular}
\end{small}
\end{center}
\vskip -0.1in
\end{table}

\begin{figure}
\centering
    \subfigure[]{\includegraphics[height=3.5cm,trim=0 0 0 0,clip]{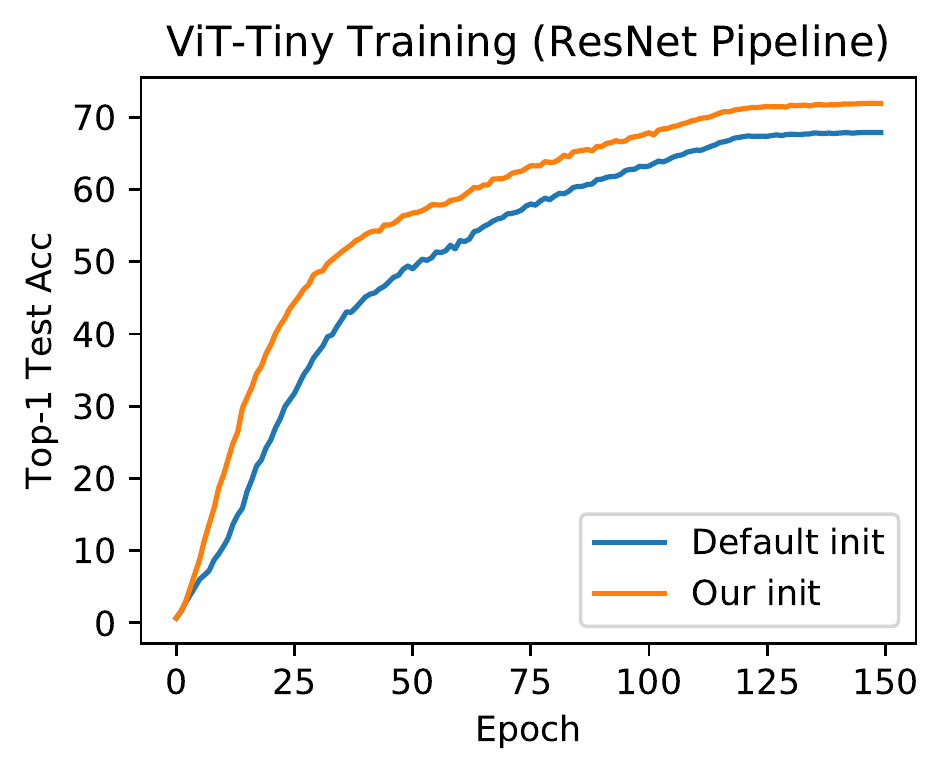}}
    \subfigure[]{\includegraphics[height=3.5cm,trim=40 0 0 0,clip]{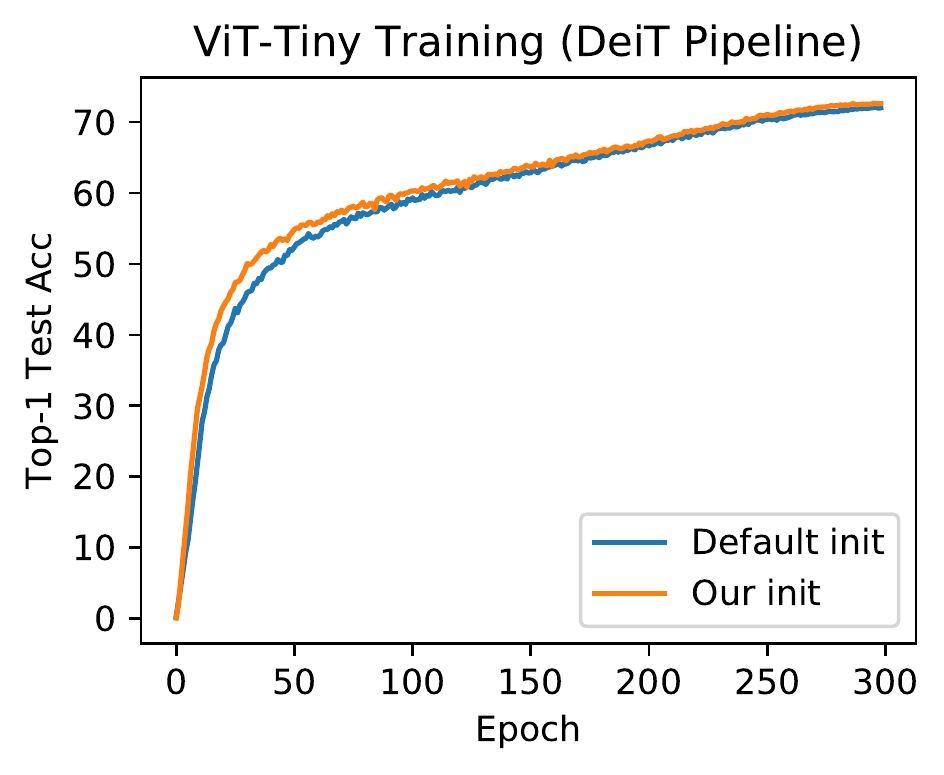}}
\vspace*{-0.5em}
    \caption{Training curves for DeiT-Tiny in a (a) ResNet-style training pipeline and a (b) DeiT-style pipeline. In the ResNet pipeline, we see a 4.1\% improvement,
    compared to a 0.5\% improvement in the DeiT pipeline.}
\label{fig:imnet-curves}
\end{figure}

\paragraph{Internal resolution}
ViTs are typically trained using high-resolution inputs
and large patch sizes.
In contrast, we trained
most of our models on CIFAR-10 using small $32\times32$
inputs and $2\times2$ patches.
Consequently, we investigate how the choice of
patch and input size affects performance.
In Appendix~\ref{sec:additional}, Table~\ref{table:internal-res},
we can see that our initialization is quite beneficial for many such
combinations.

\begin{figure}
    \centering
    \includegraphics[height=3.5cm,trim=0 0 0 0,clip]{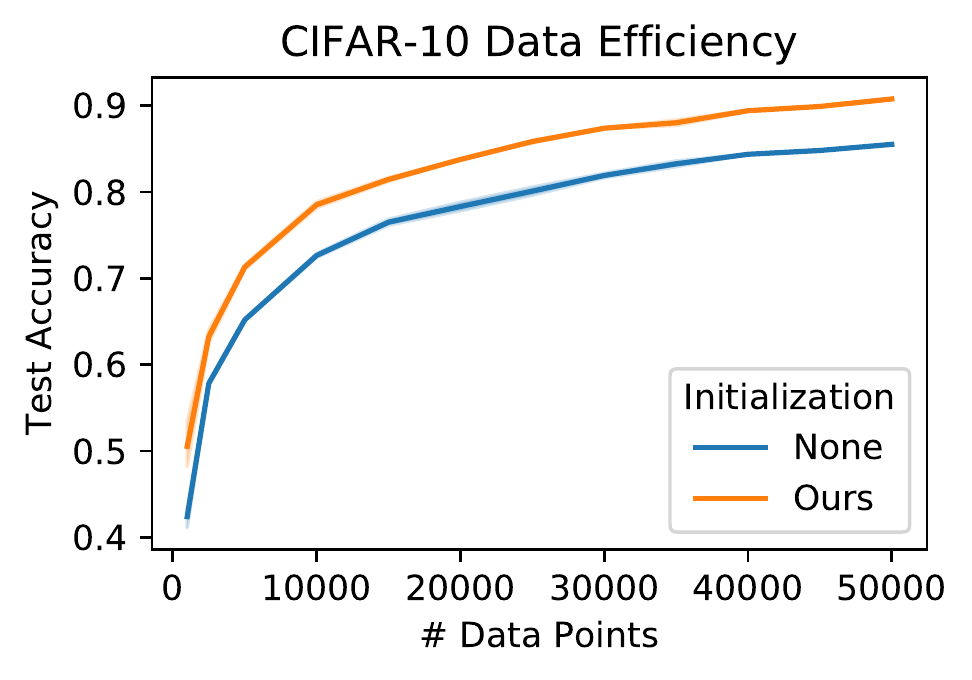}
    \caption{Adjusting the number of training points on CIFAR-10.}
    \label{fig:data-eff}
\end{figure}

\paragraph{Data efficiency}
We hypothesize that our initialization leads ViTs
to have an inductive bias more suitable for images,
and thus would expect the initialiation to be associated with especially high performance
gains on small datasets.
Consequently, we trained on a variety of subsets of CIFAR-10 (see Fig.~\ref{fig:data-eff}).
Surprisingly, we did not see performance gains inversely proportional to the size
of the dataset.
More research, \emph{e.g.,} on larger datasets, would be necessary to understand
how our initialization changes the data requirements of ViTs.

\paragraph{Other Transformer initializations}
While the motivation of our initialization is substantially different from
that of other Transformer initialization techniques, we provide
some comparisons in Table~\ref{table:other-inits}.
T-Fixup~\citep{tfixup} and ZerO~\citep{zero} focus on initializing the whole
network rather than just the self-attention layers.
For ZerO initialization, we only apply the initialization to self-attention layers.
For T-Fixup, we apply the initialization to both self-attention and MLPs.
Nonetheless, T-Fixup harms performance relative to the baseline,
and ZerO offers only a small improvement.

\paragraph{Tuning hyperparameters}
It is infeasible for us to search over all combinations
of $\alpha_{i}$ and $\beta_{i}$,
so we first fixed $\alpha_1$ and $\beta_1$ according to a guess of (0.6, 0.3),
and then tuned $\alpha_2$ and $\beta_2$.
From this, we chose $\alpha_2, \beta_2$.
Then, holding this fixed, we tuned $\alpha_1, \beta_1$.
Our grid search was performed for 100-epoch CIFAR-10 training
on a ViT-Tiny. We visualize this search in Appendix~\ref{sec:additional}, Fig.~\ref{fig:tuning}.

\subsection{ImageNet}

Here, we show that our initialization benefits training ViTs from
scratch on another relatively ``small'' dataset (for Transformers): ImageNet-1k.
We test two settings: a ResNet-style~\cite{wightman} training pipeline with 150 epochs
and standard cross-entropy loss (\emph{i.e.,} the technique of \citet{convmixer}),
and the 300-epoch DeiT training pipeline from~\citet{deit}.
In both cases, we see significant improvements for using our initialization,
with gains between 2.8-4.1\% for a ViT-Tiny in the ResNet-style pipeline
and around 0.5\% in the DeiT pipeline.
We find it surprising that we see relatively high gains even for very-long training times.
Notably, we used the same hyperparameters as found for the CIFAR-10 experiments,
though with a position embedding scale of 1.

The large performance in the ResNet-style training pipeline is particularly notable.
One of the main contributions of \citet{deit} was to propose a particular training pipeline
which was effective for training ViTs on ImageNet-scale datasets,
as ViTs did not work well in ResNet-style training pipelines.
However, our initialization provides a major boost in accuracy for ViT-Tiny in this setting,
suggesting that it begins to bridge the gap between ViT and ResNet training.

In Fig.~\ref{fig:imnet-curves}, we show training progress for both ViT training pipelines;
the difference is smaller for the DeiT pipeline, which has a larger batch size and more epochs.

\subsection{Other Datasets}
To further show that our initialization is not overfit to CIFAR-10 or ImageNet in particular,
we present results for CIFAR-100, SVHN, and Tiny ImageNet using our initialization.
We use the same settings as before with a ViT-Tiny, though with $4 \times 4$ patches
for TinyImageNet.
In Table~\ref{table:cifar100}, we see that our initialization
leads to improvements in test accuracy over 5\% for Tiny ImageNet and CIFAR-100,
but only 0.39\% for the perhaps-easier SVHN dataset.

\begin{table}
\caption{Our initialization on other datasets (ViT-Tiny, 100 epochs).}
\label{table:cifar100}
\vspace{-0.5em}
\label{table:cifar}
\vskip 0.15in
\begin{center}
\begin{small}
\begin{tabular}{lccr}
\toprule
\thead{Dataset} & \thead{Acc.\\(Base)} & \thead{Acc.\\(Init)} & \thead{$\Delta$ Acc.} \\
\midrule
Tiny ImageNet & 45.24 & 50.87 & 5.63 \\
CIFAR-100 & 60.94 & 67.33 & 6.39 \\
SVHN & 96.40 & 96.79 & 0.39 \\
\bottomrule
\end{tabular}
\end{small}
\end{center}
\vskip -0.1in
\end{table}

\section{Why does this initialization work?}
\label{sec:why-works}

We have shown that our mimetic initialization is quite effective for enhancing
visual recognition on small datasets. 
Here, we propose some additional explanations
for why our method is effective.
The first section concerns the query and key weights,
while the next two investigate the somewhat-more-mysterious
negative diagonal of the value and projection product.

\paragraph{Near-identity attention maps.}
In Fig.~\ref{ref:attn-maps} and Eq.~\ref{eq:near-eye}, we see that our initialization,
much like pretraining,
makes the attention maps somewhat similar to identity matrices, particularly
in earlier layers.
The resemblance of our attention maps using our initialization to those in pretrained
models is notable in itself.
\citet{dks} notes that forcing attention maps to be the identity
avoids rank collapse, which can otherwise prevent trainability.
However, they note that exact-identity attention cannot pass
gradients to the query and key parameters, meaning it is not actually a viable
initialization technique.
We hypothesize that our initialization strikes a balance between
untrained attention maps (as in Fig.~\ref{ref:attn-maps}a)
and identity attention maps.

\paragraph{LayerScale analogy}
In \citet{layerscale}, a simple technique called LayerScale
is proposed to train deeper Transformers more effectively,
in which the layer at a skip connection is multiplied
by a learnable diagonal scaling matrix $D$:
\begin{equation}
	X_l' = X_l + D \cdot \mathsf{SelfAttn}(\eta(X_l))
\end{equation}
where $\eta$ denotes LayerNorm.
Here, we show that the way we initialize
$W_V, W_{proj}$ has a relatively weak resemblance to this technique.
Considering Eq.~\ref{eq:near-eye}, we approximate
the attention maps after our initialization as being close
to the identity, and assume that $\eta(X_l) \approx X_l$:
\begin{align}
	X_l' &= X_l + \mathsf{SelfAttn}(\eta(X_l)) \\
		 &\approx X_l + I \eta(X_l) W_V W_{proj} \\
		 &\approx X_l + I \eta(X_l) (\alpha Z - \beta I) \quad \text{(due to our init)} \\
		 &\approx (I - \beta) X_l + \alpha \eta(X_l) Z 
\end{align}

Scaling $X_l$ by $(I - \beta)$ is similar in spirit to LayerScale,
except in our case we are multiplying the left-hand instead of the right-hand
term in the skip connection.
This motivates us to compare our technique for setting to $W_V W_{proj}$ to using
LayerScale, or our variant of LayerScale above.

\begin{table}
\caption{Other initializations}
\label{table:other-inits}
\vskip 0.15in
\begin{center}
\begin{small}
\begin{tabular}{ccccc}
\toprule
\thead{T-Fixup} & \thead{ZerO} & \thead{LayerScale \\ Original} & \thead{LayerScale \\ Our Version}  & \thead{Our \\ Initialization} \\
\midrule
85.38 & 87.41 & 89.90 & 88.68 & 91.38 \\
\bottomrule
\end{tabular}
\end{small}
\end{center}
\end{table}

We searched ten choices of initialization for the diagonal elements in $[0, 1]$ for both LayerScale
techniques, replacing our $W_V W_{proj}$ initialization, and report the best results in Table \ref{table:other-inits}.
Note we leave our $W_Q W_K^T$ initialization unchanged.
Neither method achieves the performance of ours (with a difference of about 1.5\%)
though LayerScale comes closest.
We conclude that the benefits of our initialization extend beyond
its possible similarity to LayerScale.

\paragraph{Convolution analogy.}

Many works which successfuly train ViTs on small datasets
do so by adding aspects of convolution, whether implicitly or explicitly.
Here, we explore adding locality to self-attention through convolutional biases:
\begin{equation}
    \mathsf{Softmax} \left( X W_Q W_K^T X^T + \gamma C \right),
\end{equation}
where $C$ is a doubly-block circulant convolution matrix and $\gamma$ is a learnable scalar.
Here, $C$ is reminiscent of the $PP^T$ term in Eq.~\ref{eq:near-eye}.
This achieves 87.5\% accuracy on CIFAR-10 within our usual training pipeline (without our init).
For comparison, plain self-attention with no special initialization achieves 88.1\% accuracy.
Next, we move the convolution outside the softmax:
\begin{equation}
	\mathsf{Softmax} \left( X W_Q W_K^T X^T \right) + \gamma C,
\end{equation}
This has a more considerable advantange, resulting in 89.9\% accuracy.
Then, if we instead use $C' = \mathsf{Softmax}(\gamma C)$ to restrict $C'$ to be
all-positive, we achieve 75\% accuracy.
That is, it appears that the negative component of the convolution matrix is necessary.

Thus, we hypothesize that initializing $W_V W_{proj}$ to have a negative diagonal
is perhaps beneficial for the same reason: this allows for some degree of ``negative'' or edge-detector-like
spatial mixing to occur, a potentially useful starting point for the purpose of visual recognition.

\begin{figure*}
\centering
\subfigure[$W_Q W_K^T$ has a wider array of diagonal magnitudes (first 3 heads shown). $\rightarrow$ Layers 1-12, $\downarrow$ Attention Heads 1-3 of 12 ]{\includegraphics[width=\textwidth,trim=0 0 0 0,clip]{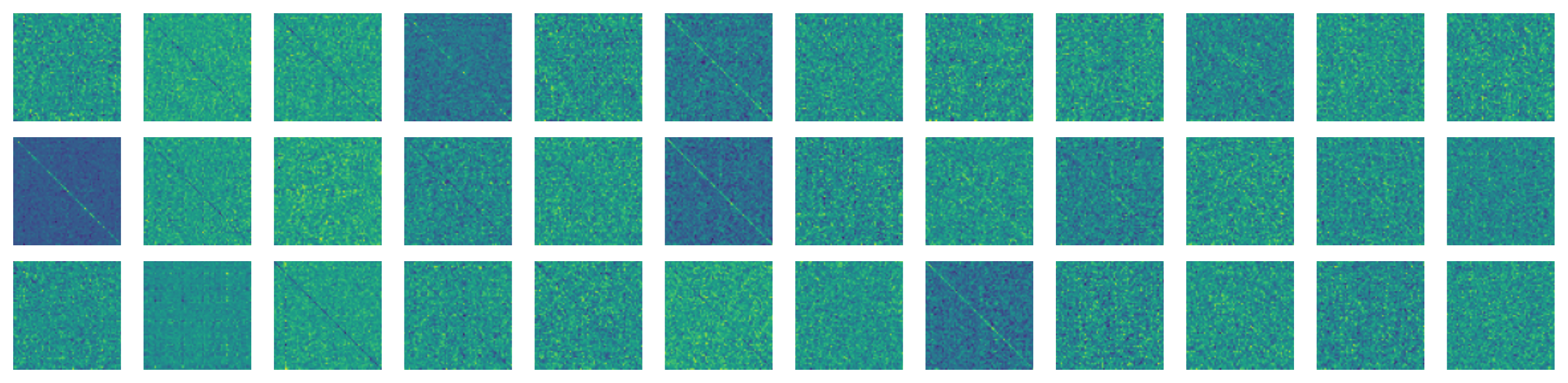}}
    \subfigure[$W_V W_{proj}$ becomes positive in deeper layers.]{\includegraphics[width=\textwidth,trim=0 0 0 0,clip]{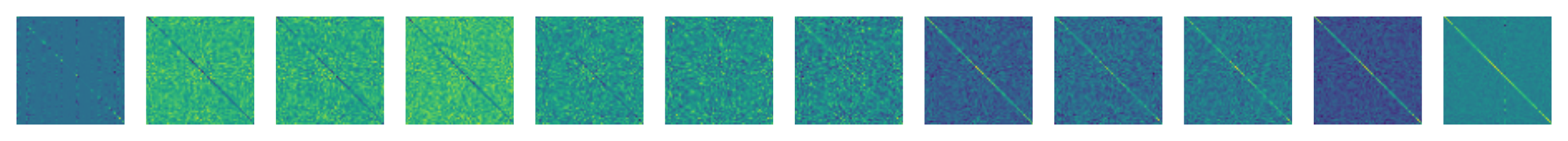}}
\vspace*{-0.5em}
\caption{A pretrained GPT-2 shows considerably different patterns in the products of $W_Q W_K^T$ and $W_V W_{proj}$, compared to ViTs.}
\label{ref:tmr-observations}
\end{figure*}

\section{Language Modeling}

While our method was primarily inspired by pretrained Vision Transformers,
in this section we investigate its potential for use in language models.
As noted in Sec.~\ref{sec:observations} and seen in Fig.~\ref{ref:tmr-observations},
we do not see precisely the same pattern in a pre-trained GPT-2 model as we do in a ViT.
Nonetheless, we use the same technique here without modification;
we saw no improvement from, \emph{e.g.,} attempting to model the positive diagonals
of $W_V W_{proj}$.

\paragraph{Small-scale}
Generally, it is hard to train Transformers from scratch on small language tasks~\cite{transformer-xl};
it requires substantial regularization, \emph{e.g.,} in the form of dropout.
For word-level modeling on Penn TreeBank (PTB), we thus add one regularization tweak:
word-level embedding dropout (\emph{i.e.,} dropout of entire embedding vectors).
This allows us to achieve sub-100 perplexity.

We use a training setup identical to that of \citet{bai}, training for 100 epochs
and reducing the learning rate when it plateaus.
We used a vanilla Transformer with sinusoidal position embeddings,
with embedding dimension 384, 12 layers, 8 attention heads, and weight-tied embeddings.

\looseness=-1
First, on char-level PTB we did a small-scale hyperparameter search for those $\alpha_i, \beta_i$
yielding the best validation BPC. We chose $\alpha_1 = 0$, $\beta_1 = 0.5$, and $\alpha_2 = \beta_2 = 0.2$.
We used these parameters on subsequent word-level modeling tasks.
On char-level PTB, we see a small but significant reduction in BPC from 1.233 to 1.210
through using our initialization.
Similarly, we see a small reduction in perplexity on word-level PTB, from 84.84 to 82.34. 
(For both tasks, smaller is better.)

While our initialization does not make a large amount of difference for these small-scale language
tasks as it does for vision tasks, it does show a small amount of improvement.
We suspect that it may be the case that a mimetic initialization scheme more finely-tuned
to the language setting may show still better performance.

\paragraph{Medium-scale}
Next, we tried our initialization on a larger-scale task, WikiText-103. 
Here, we used an embedding dimension of 410 with 16 layers, 10 heads,
and sinusoidal embeddings, with the same hyperparameters as for the previous task.
As this dataset is around 110 times larger than PTB, we trained for only 50 epochs.
Here, we see a more significant performance gain from using our initialization,
reducing the test perplexity from 28.87 to 28.21 (see Table~\ref{table:language}).
While this is not a massive improvement, this is consistent with our observation
on vision tasks that the improvement from our technique may be more significant for larger models.
Further, we also note that in this case the number of parameters being initialized is quite small relative to the total
number of parameters of the language model due to the word embedding weights,
something which does not occur with vision models.

\begin{table}[t]
\caption{Language results}
\label{table:language}
\vskip 0.15in
\begin{center}
\begin{small}
\begin{tabular}{lccc}
\toprule
\thead{Task} & \thead{Metric} & \thead{Base} & \thead{Init} \\
\midrule
Char-level PTB & bpc & 1.233 & \textbf{1.210} \\
Word-level PTB & ppl & 84.84 & \textbf{82.34} \\
WikiText-103 & ppl & 28.87 & \textbf{28.21} \\
\bottomrule
\end{tabular}
\end{small}
\end{center}
\vskip -0.1in
\end{table}

\section{Conclusion}
Our proposed initialization technique for Transformers is particularly effective at improving performance
on small-scale image recognition tasks, leading to an increase of over 5\% accuracy in some cases.
In other words, we address the problem that Vision Transformers are hard to train in ResNet-style pipelines
solely through a structured initialization of the weights, without need for any kind of pretraining or architectural modifications.
To a lesser extent, we demonstrated that our initialization leads to non-trivial gains on WikiText-103, showing that it
also has potential to similarly improve language modeling on relatively small datasets.
More broadly, we proposed a class of techniques we call \emph{mimetic initialization}, in which we attempt to gain some benefits
of pretraining by mimicking the surface-level qualities of pretrained models.
We speculate that it may be possible to use domain knowledge to ``program'' models before
training in order to reach more desirable optima that may have been out of reach with a completely random initialization.
With better structured initialization techniques like our own,
perhaps Transformers really \emph{are} the universal architecture.

\bibliography{example_paper}
\bibliographystyle{icml2023}

%%%%%%%%%%%%%%%%%%%%%%%%%%%%%%%%%%%%%%%%%%%%%%%%%%%%%%%%%%%%%%%%%%%%%%%%%%%%%%%
%%%%%%%%%%%%%%%%%%%%%%%%%%%%%%%%%%%%%%%%%%%%%%%%%%%%%%%%%%%%%%%%%%%%%%%%%%%%%%%
% APPENDIX
%%%%%%%%%%%%%%%%%%%%%%%%%%%%%%%%%%%%%%%%%%%%%%%%%%%%%%%%%%%%%%%%%%%%%%%%%%%%%%%
%%%%%%%%%%%%%%%%%%%%%%%%%%%%%%%%%%%%%%%%%%%%%%%%%%%%%%%%%%%%%%%%%%%%%%%%%%%%%%%
\newpage
\appendix
\onecolumn

\section{Additional results}
\label{sec:additional}

\begin{figure}
\centering
	\subfigure[Untrained]{\includegraphics[width=0.15\textwidth,trim=0 0 0 0,clip]{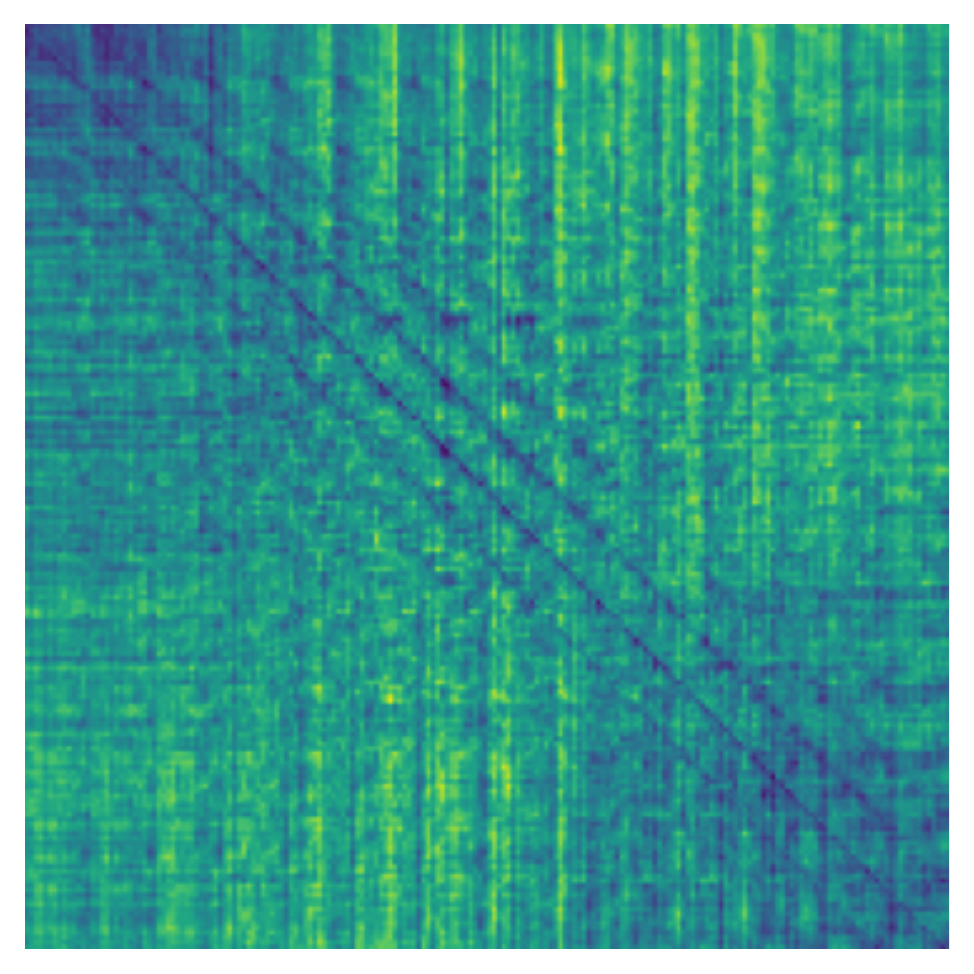}}
	\hfill
	\subfigure[CIFAR-10 trained]{\includegraphics[width=0.15\textwidth,trim=0 0 0 0,clip]{fig/attn/0-cifar-trained.pdf}}
	\hfill
	\subfigure[ImageNet pretrained]{\includegraphics[width=0.15\textwidth,trim=0 0 0 0,clip]{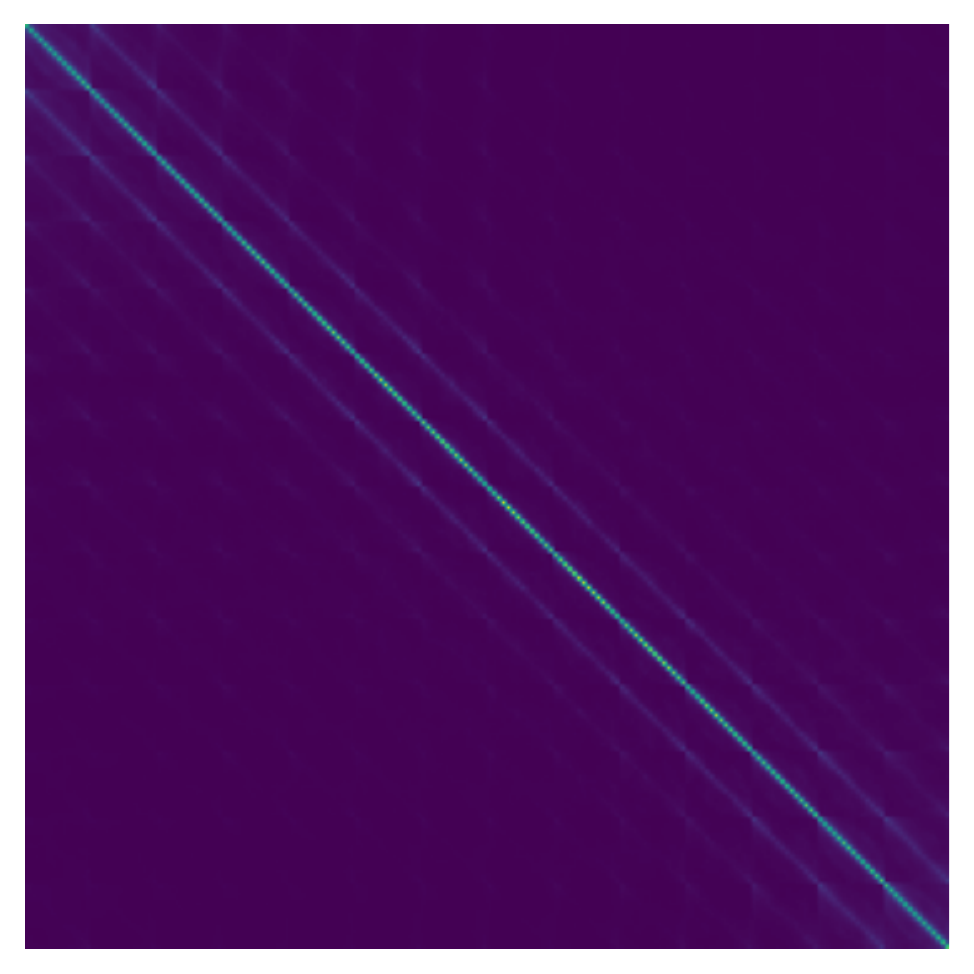}}
	\hfill
	\subfigure[Our init]{\includegraphics[width=0.15\textwidth,trim=0 0 0 0,clip]{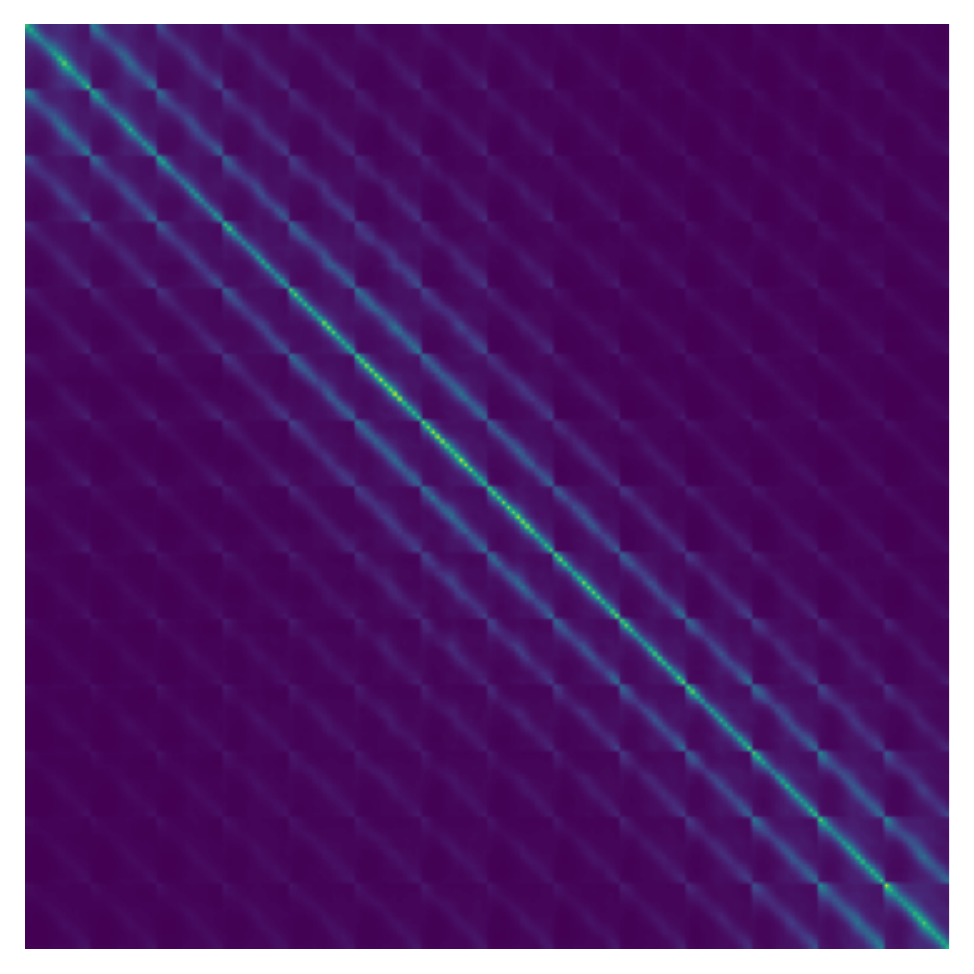}}
	\hfill
	\subfigure[Our init CIFAR-10 trained]{\includegraphics[width=0.15\textwidth,trim=0 0 0 0,clip]{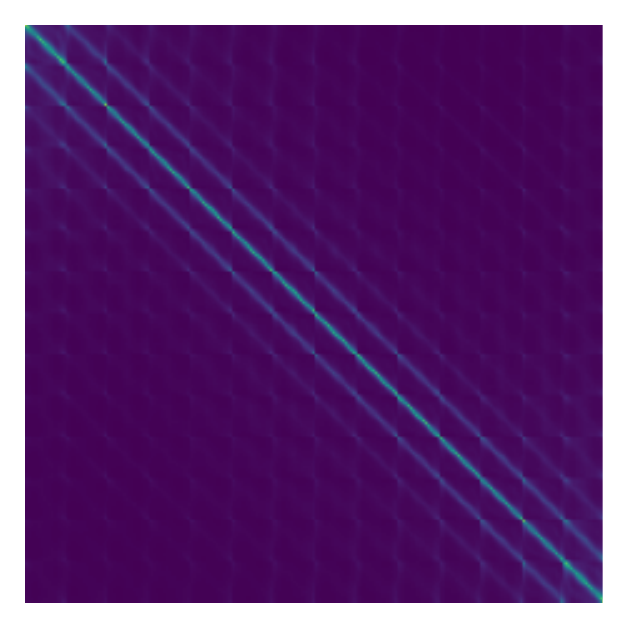}}
    \newline
	%\vspace{-0.5em}
	\subfigure{\includegraphics[width=0.15\textwidth,trim=0 0 0 0,clip]{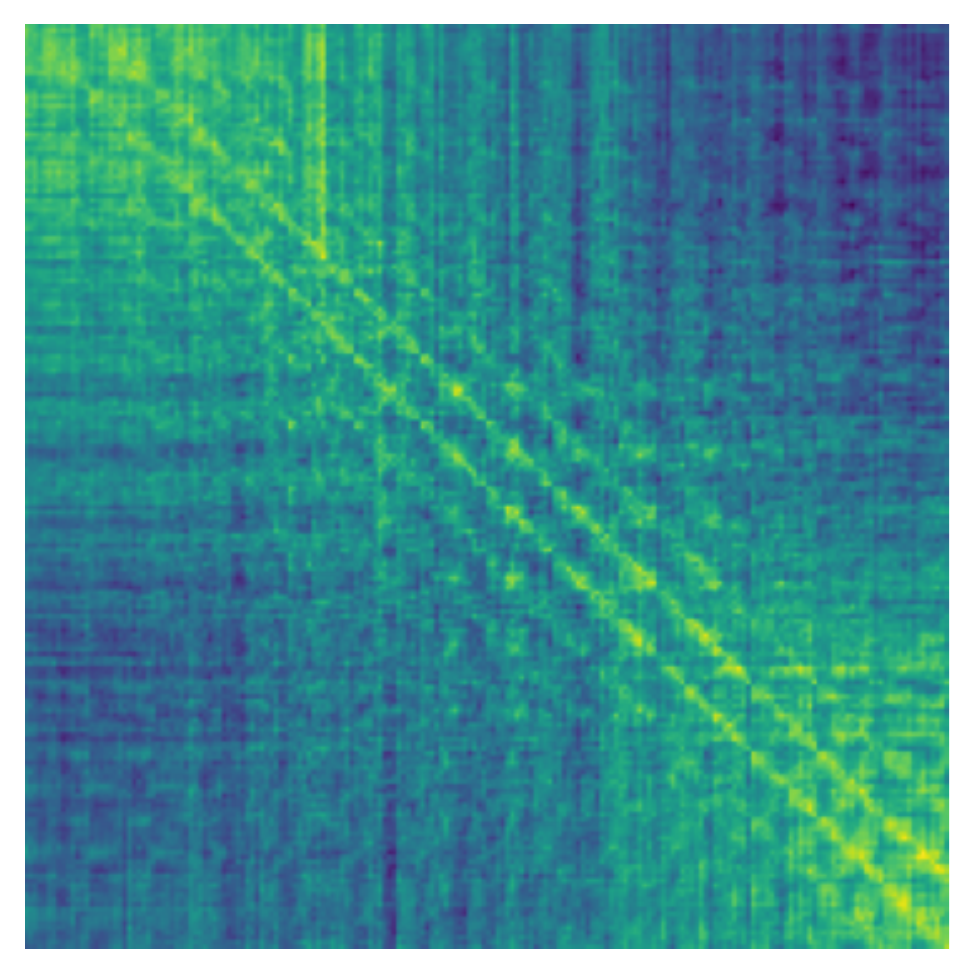}}
	\hfill
	\subfigure{\includegraphics[width=0.15\textwidth,trim=0 0 0 0,clip]{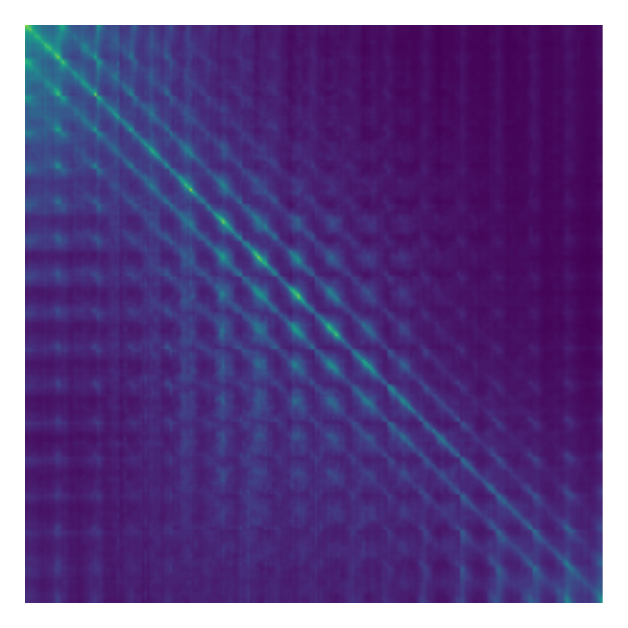}}
	\hfill
	\subfigure{\includegraphics[width=0.15\textwidth,trim=0 0 0 0,clip]{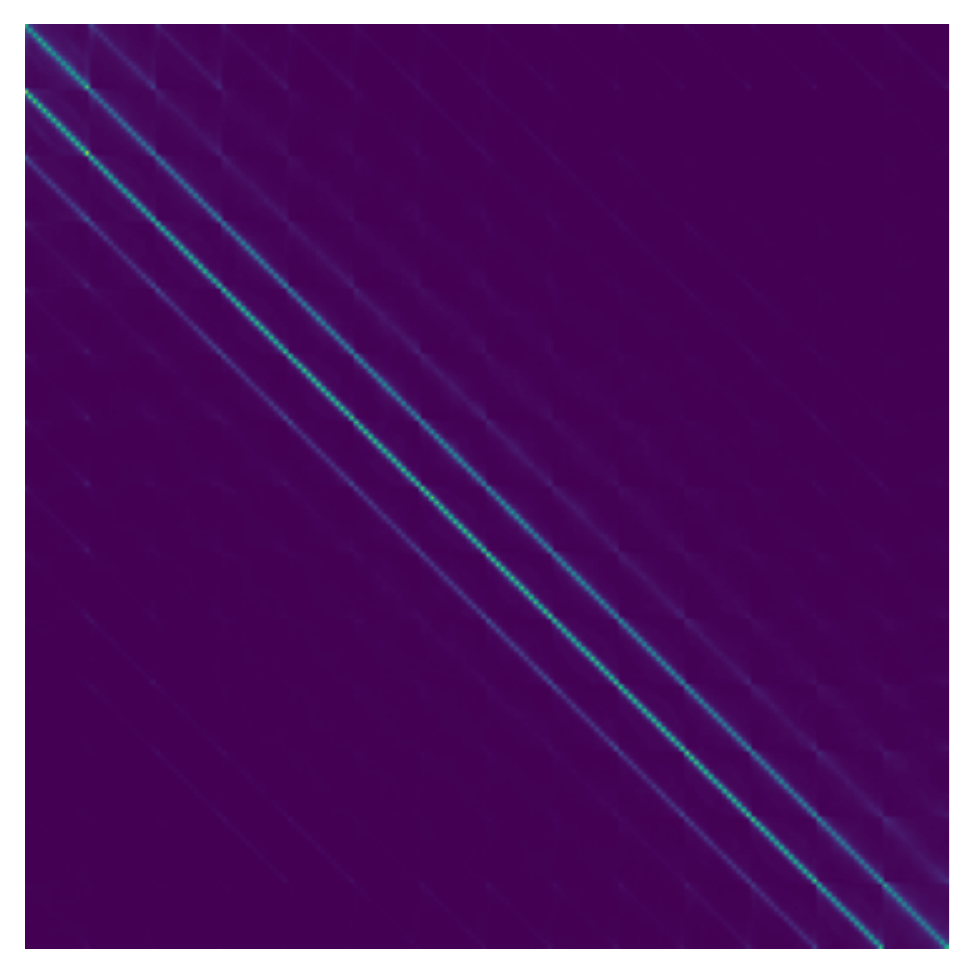}}
	\hfill
	\subfigure{\includegraphics[width=0.15\textwidth,trim=0 0 0 0,clip]{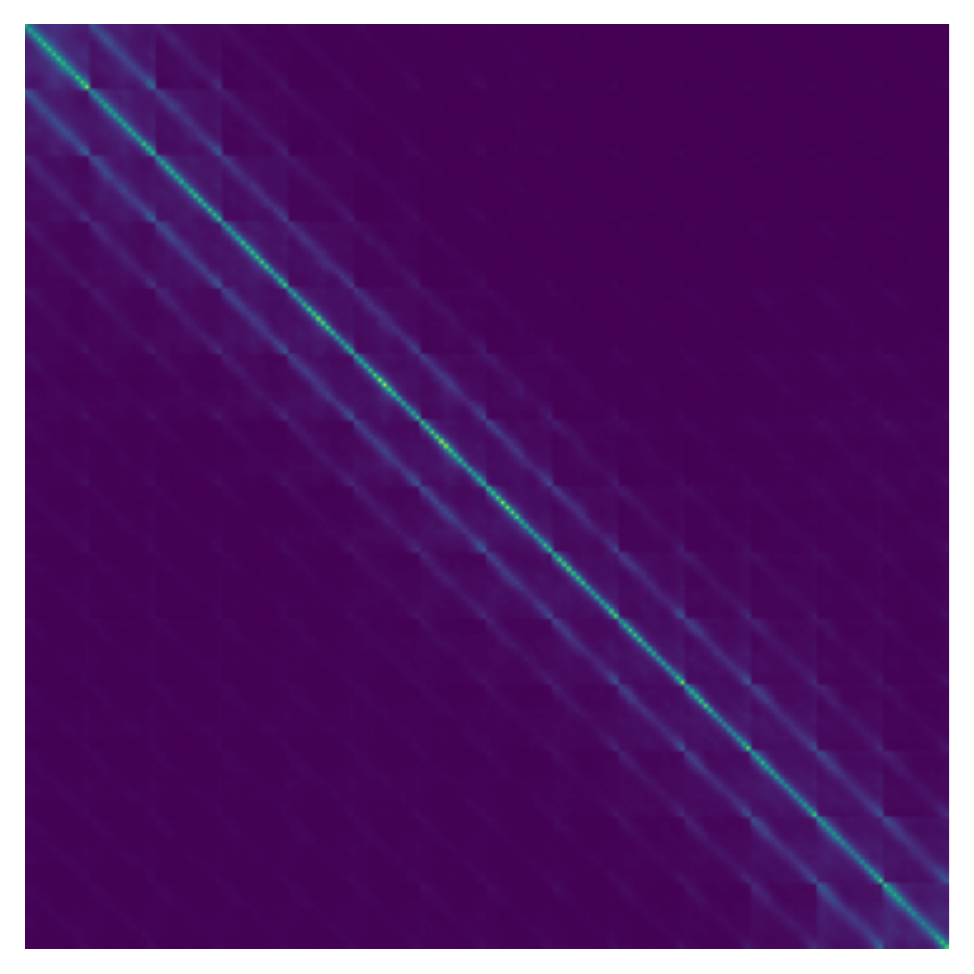}}
	\hfill
	\subfigure{\includegraphics[width=0.15\textwidth,trim=0 0 0 0,clip]{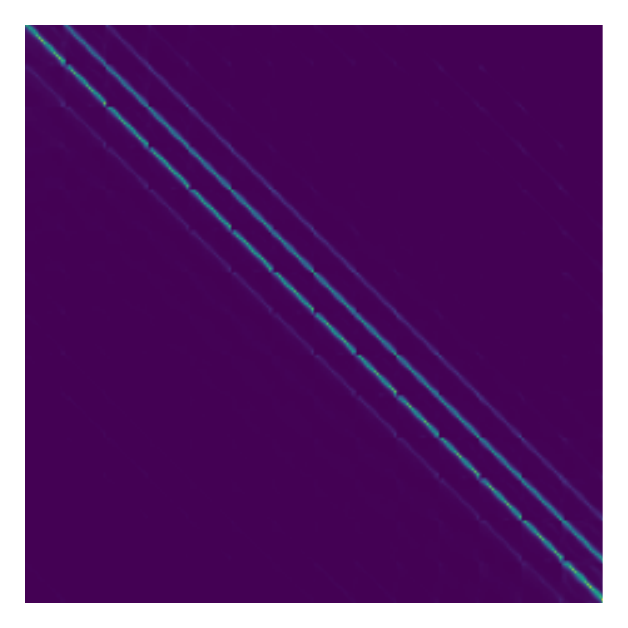}}
    \newline
	\subfigure{\includegraphics[width=0.15\textwidth,trim=0 0 0 0,clip]{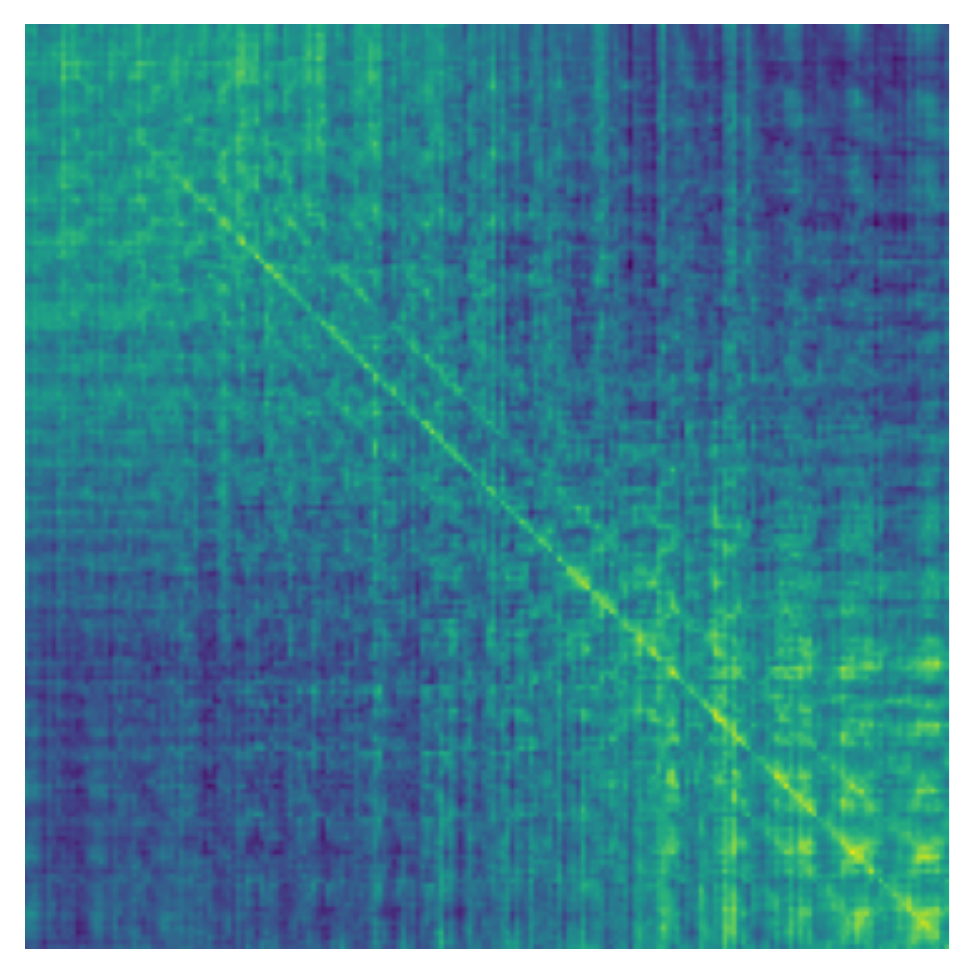}}
	\hfill
	\subfigure{\includegraphics[width=0.15\textwidth,trim=0 0 0 0,clip]{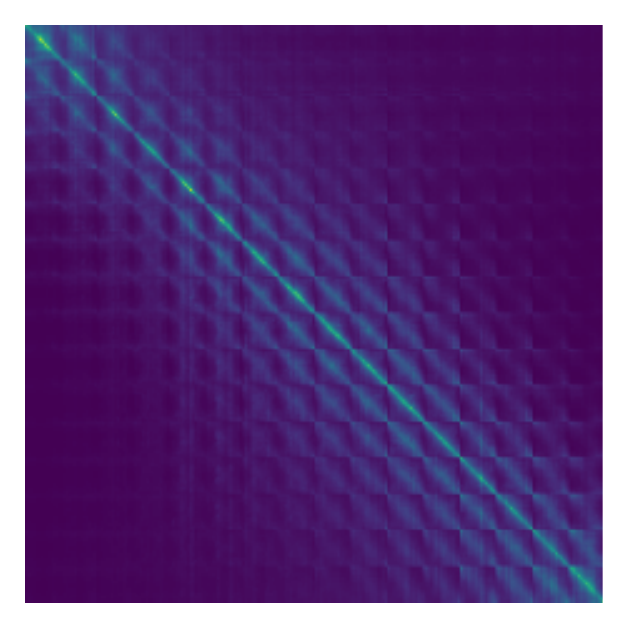}}
	\hfill
	\subfigure{\includegraphics[width=0.15\textwidth,trim=0 0 0 0,clip]{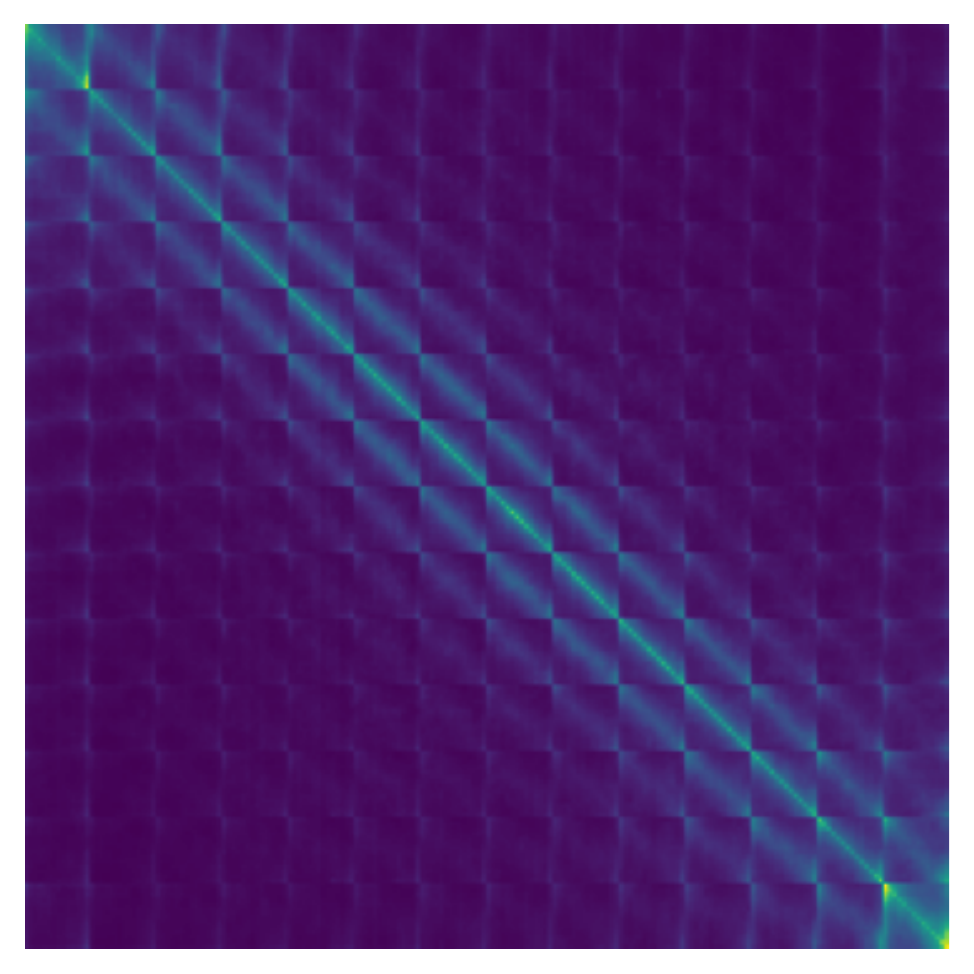}}
	\hfill
	\subfigure{\includegraphics[width=0.15\textwidth,trim=0 0 0 0,clip]{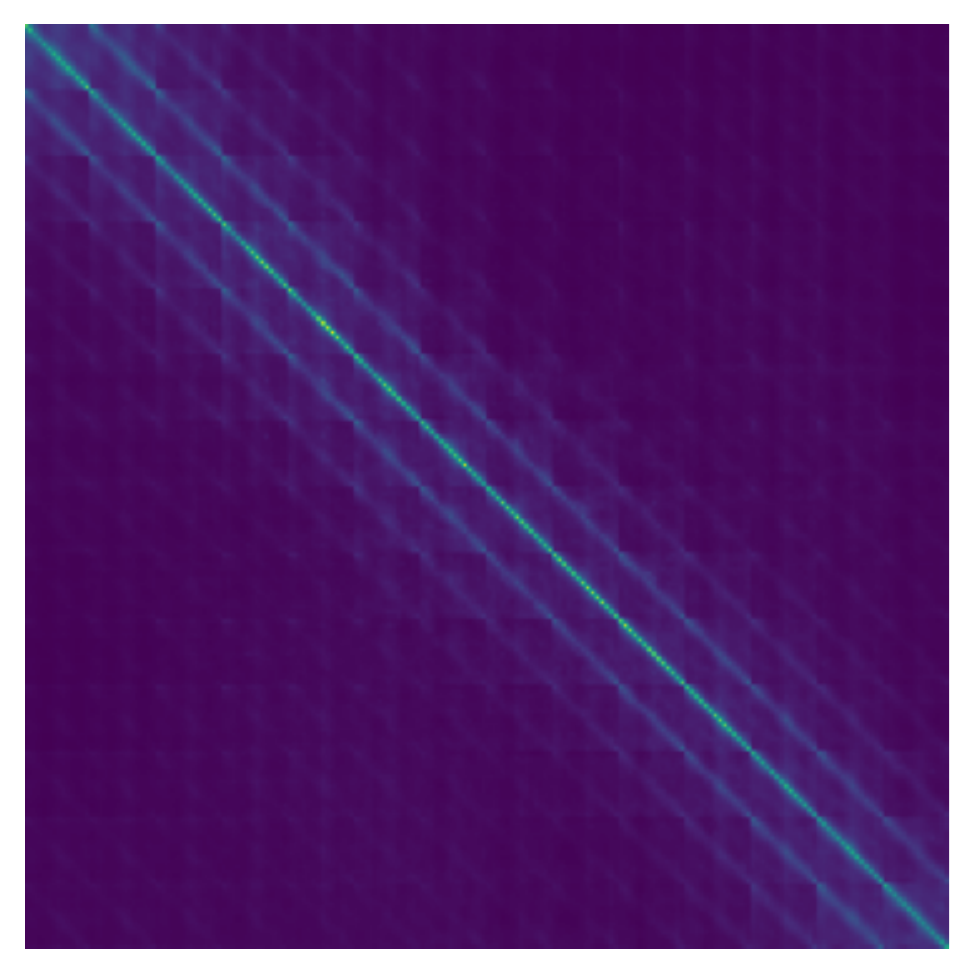}}
	\hfill
	\subfigure{\includegraphics[width=0.15\textwidth,trim=0 0 0 0,clip]{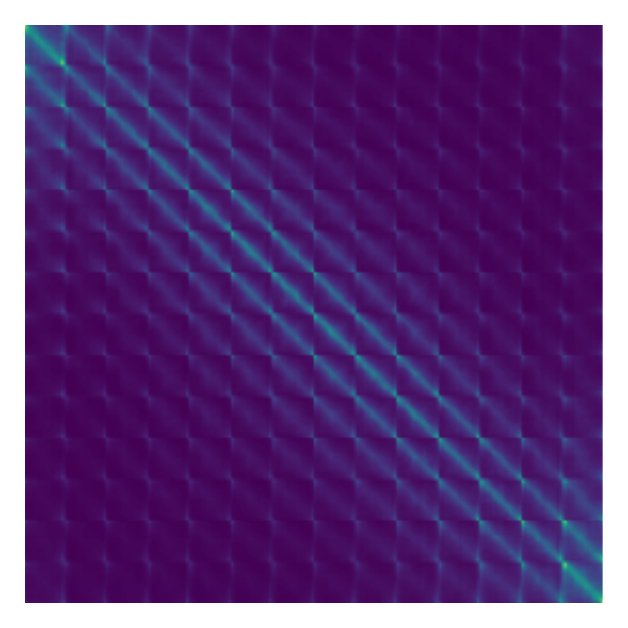}}
    \newline
	\subfigure{\includegraphics[width=0.15\textwidth,trim=0 0 0 0,clip]{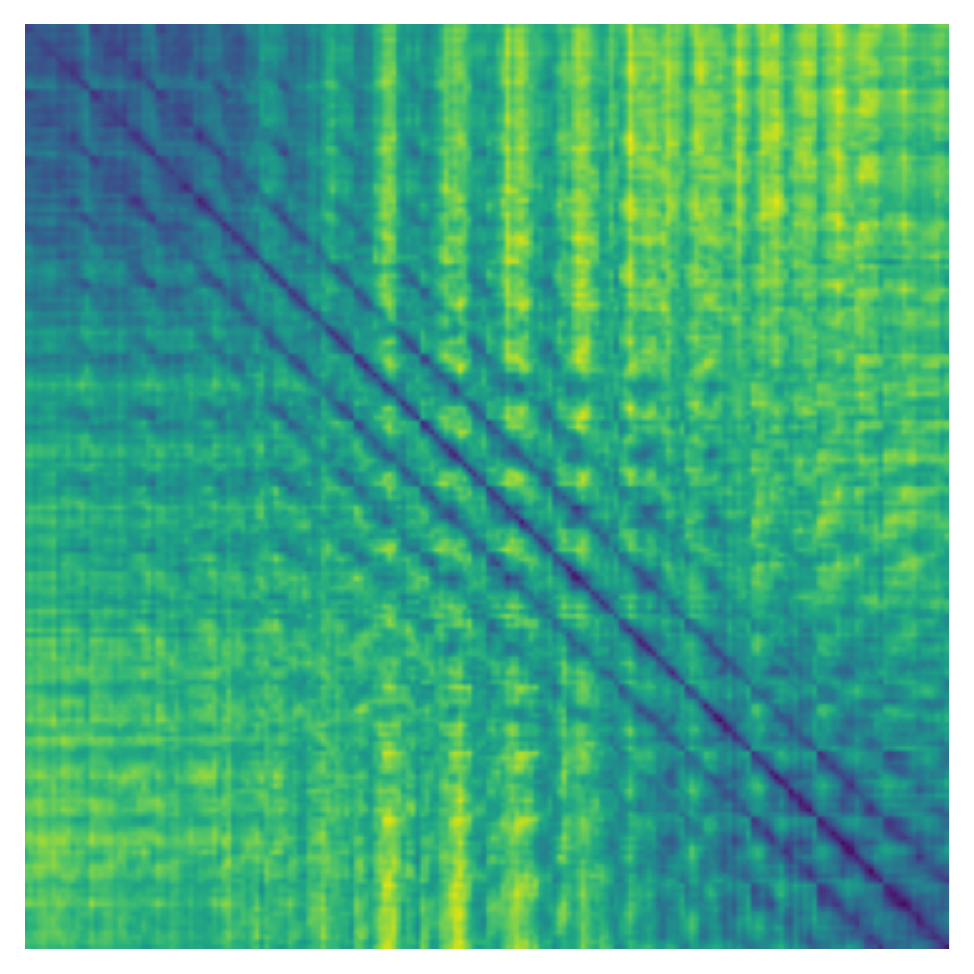}}
	\hfill
	\subfigure{\includegraphics[width=0.15\textwidth,trim=0 0 0 0,clip]{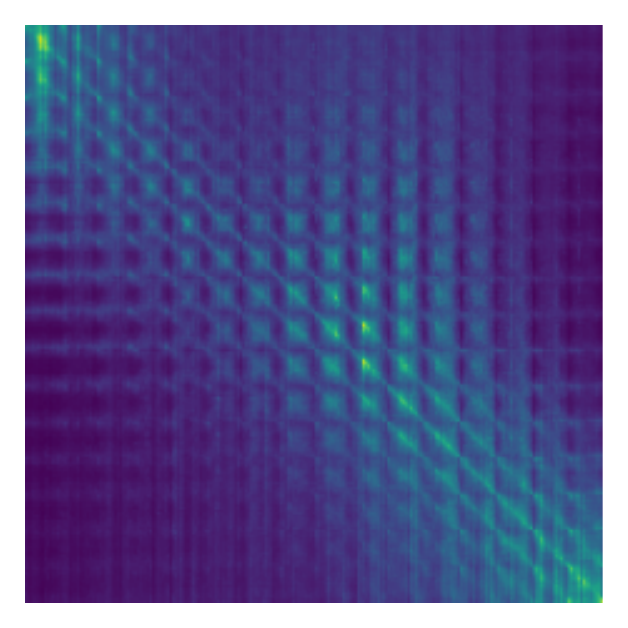}}
	\hfill
	\subfigure{\includegraphics[width=0.15\textwidth,trim=0 0 0 0,clip]{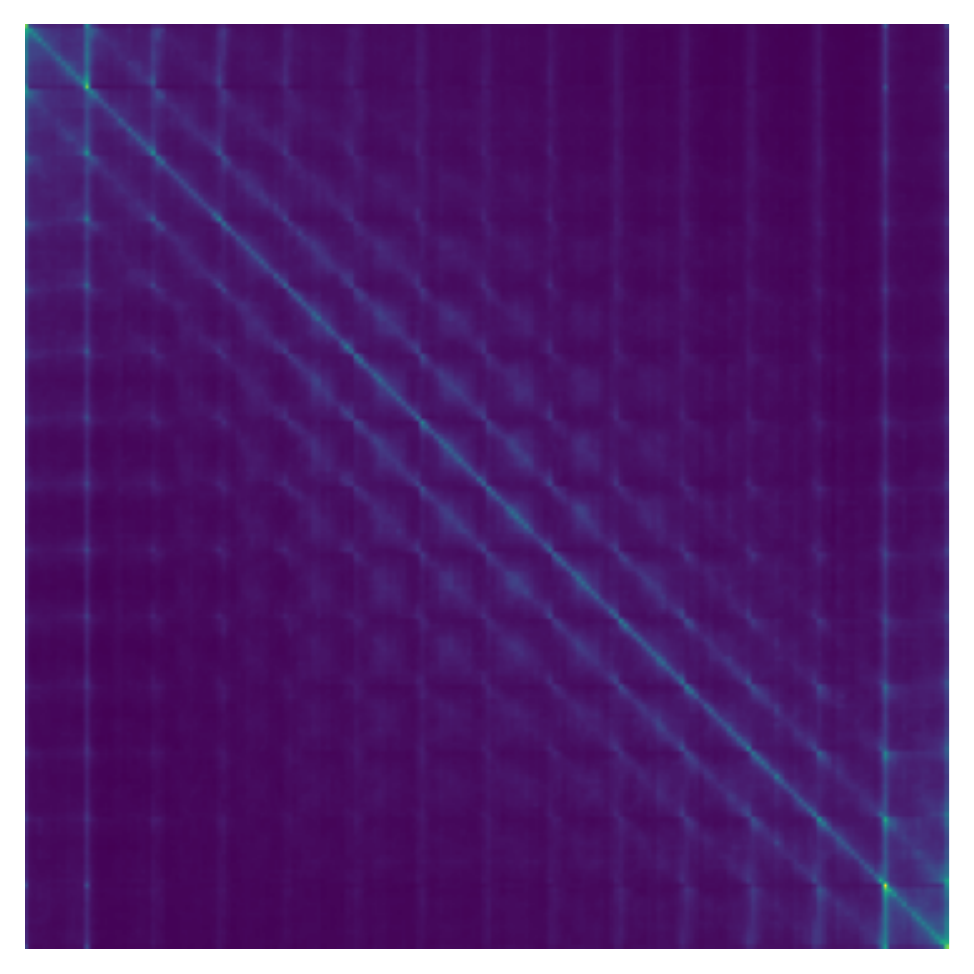}}
	\hfill
	\subfigure{\includegraphics[width=0.15\textwidth,trim=0 0 0 0,clip]{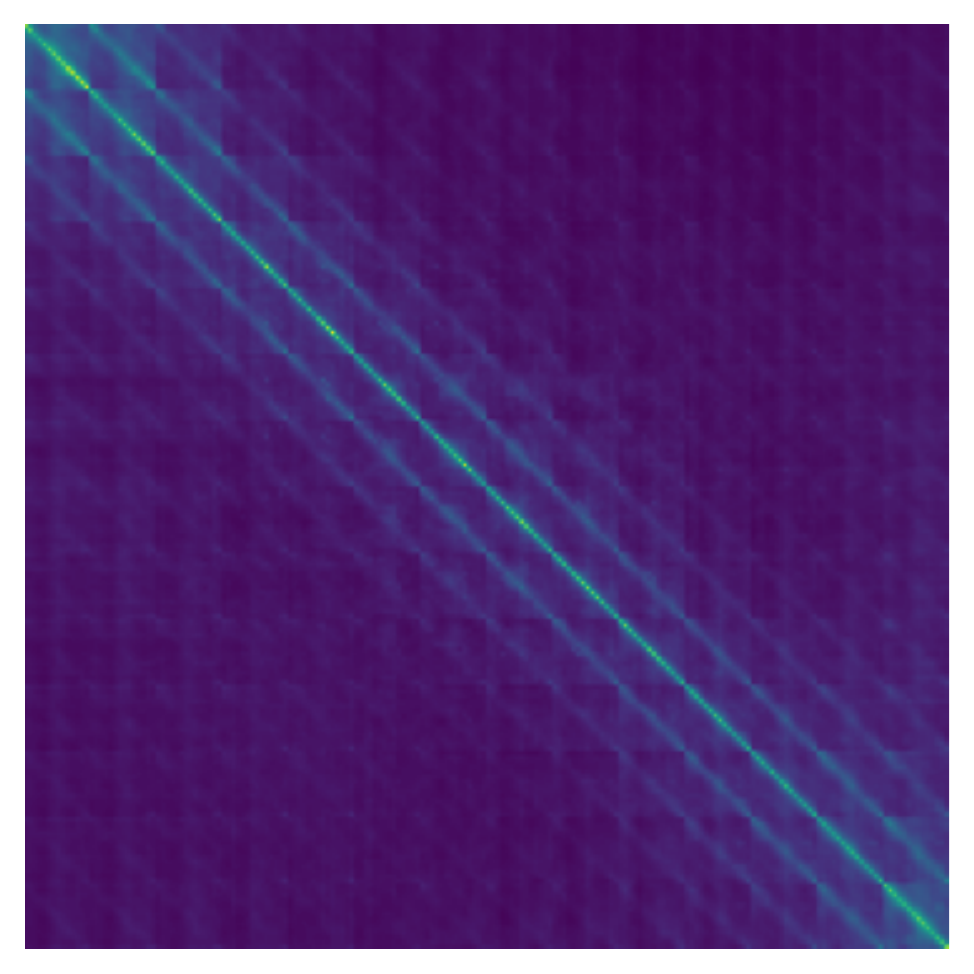}}
	\hfill
	\subfigure{\includegraphics[width=0.15\textwidth,trim=0 0 0 0,clip]{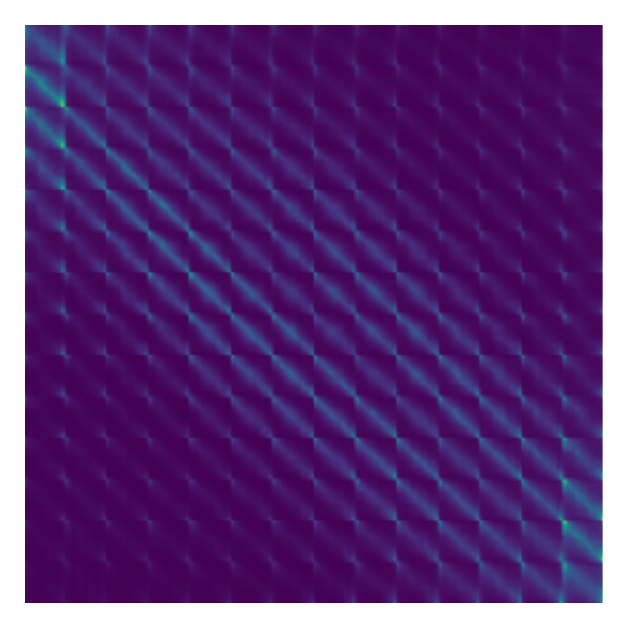}}
    \newline
	\subfigure{\includegraphics[width=0.15\textwidth,trim=0 0 0 0,clip]{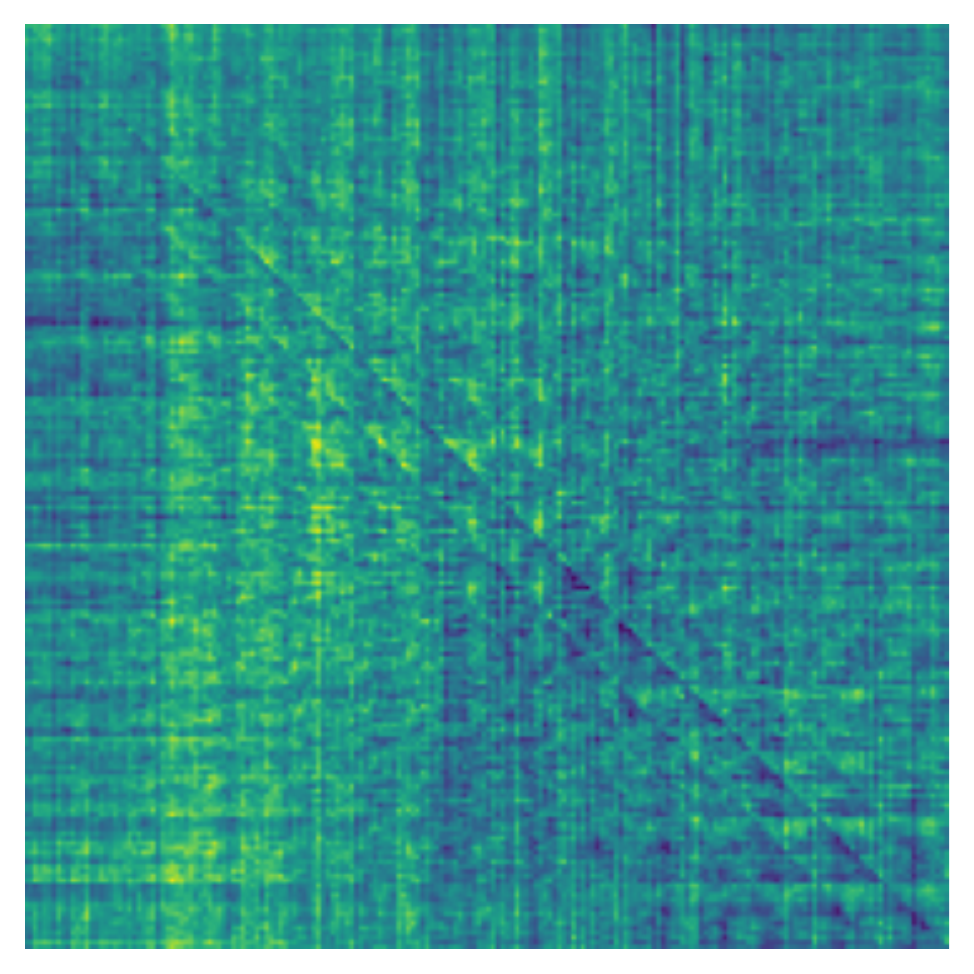}}
	\hfill
	\subfigure{\includegraphics[width=0.15\textwidth,trim=0 0 0 0,clip]{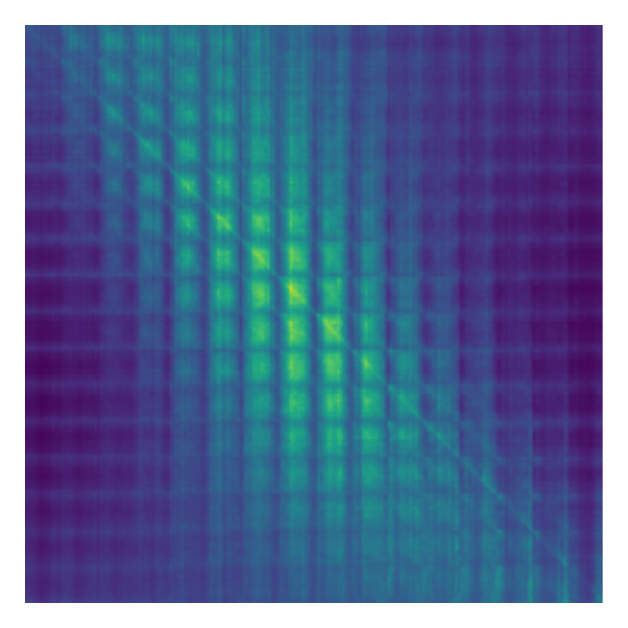}}
	\hfill
	\subfigure{\includegraphics[width=0.15\textwidth,trim=0 0 0 0,clip]{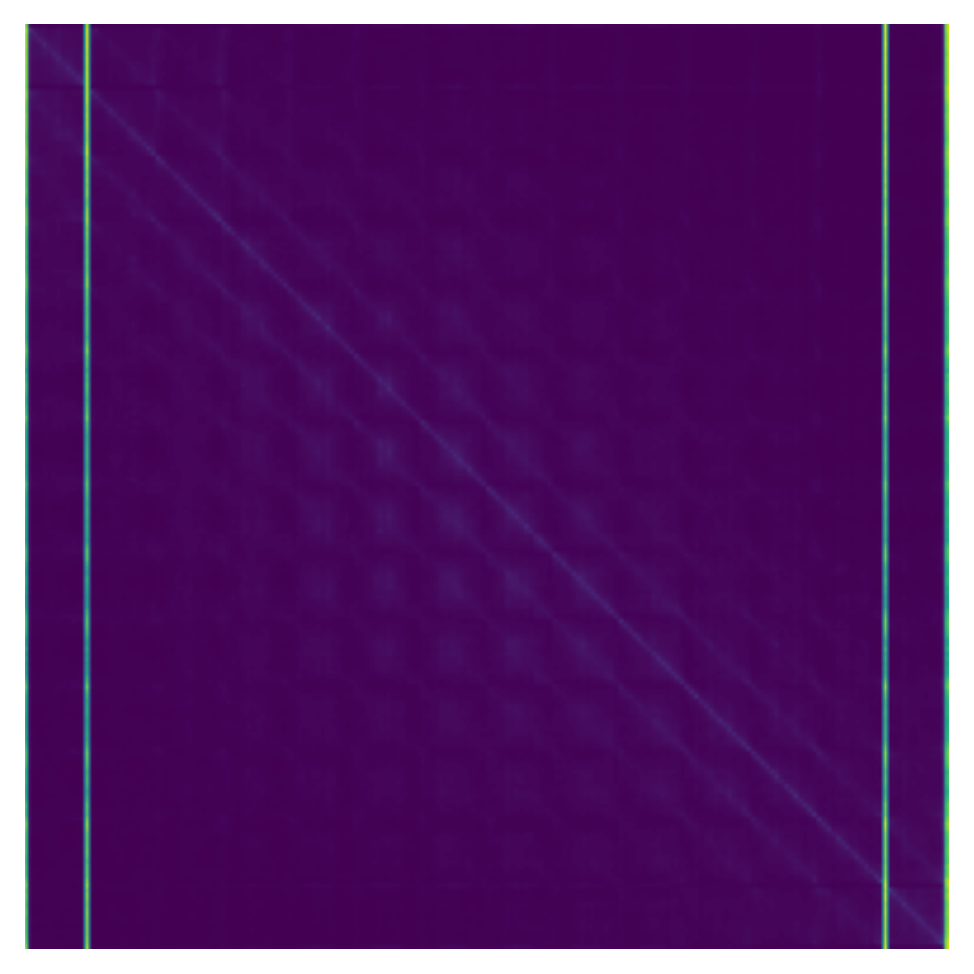}}
	\hfill
	\subfigure{\includegraphics[width=0.15\textwidth,trim=0 0 0 0,clip]{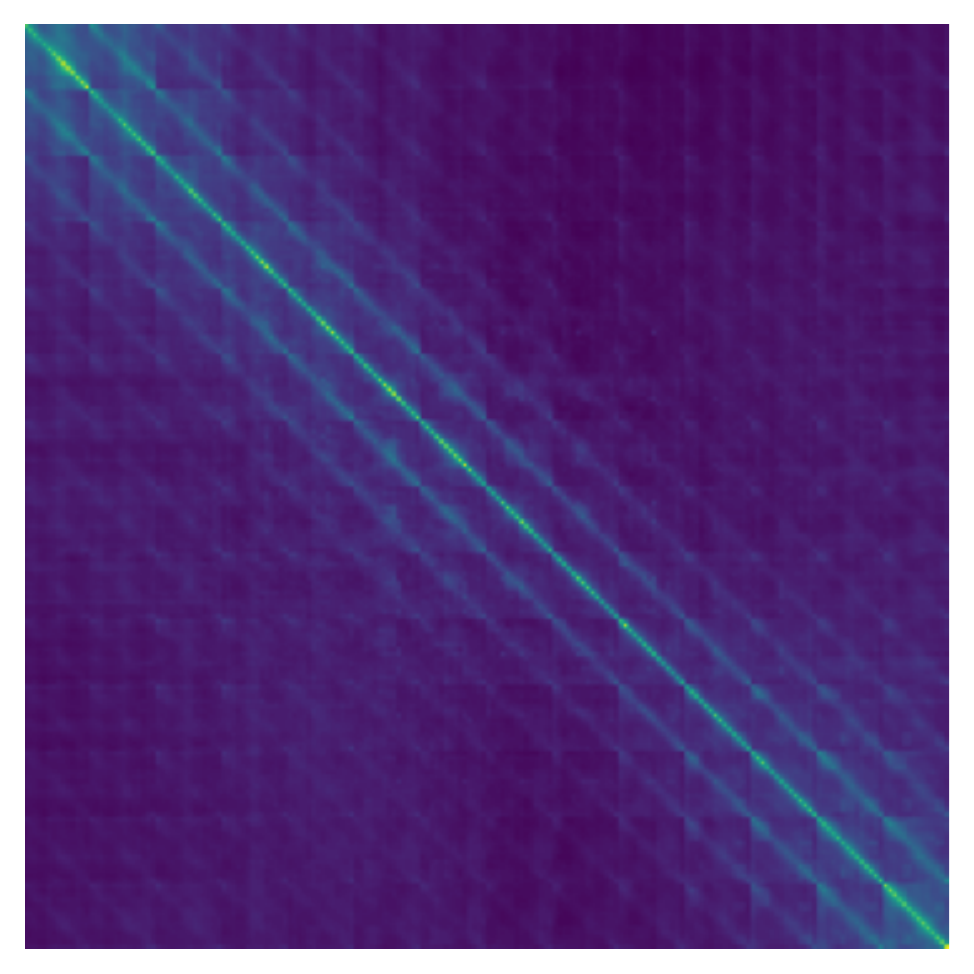}}
	\hfill
	\subfigure{\includegraphics[width=0.15\textwidth,trim=0 0 0 0,clip]{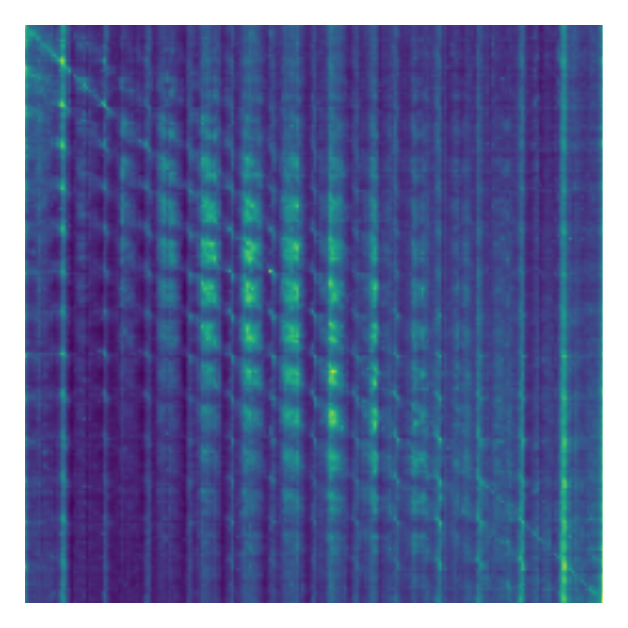}}
    \newline
	%\vspace{-0.5em}
	\subfigure{\includegraphics[width=0.15\textwidth,trim=0 0 0 0,clip]{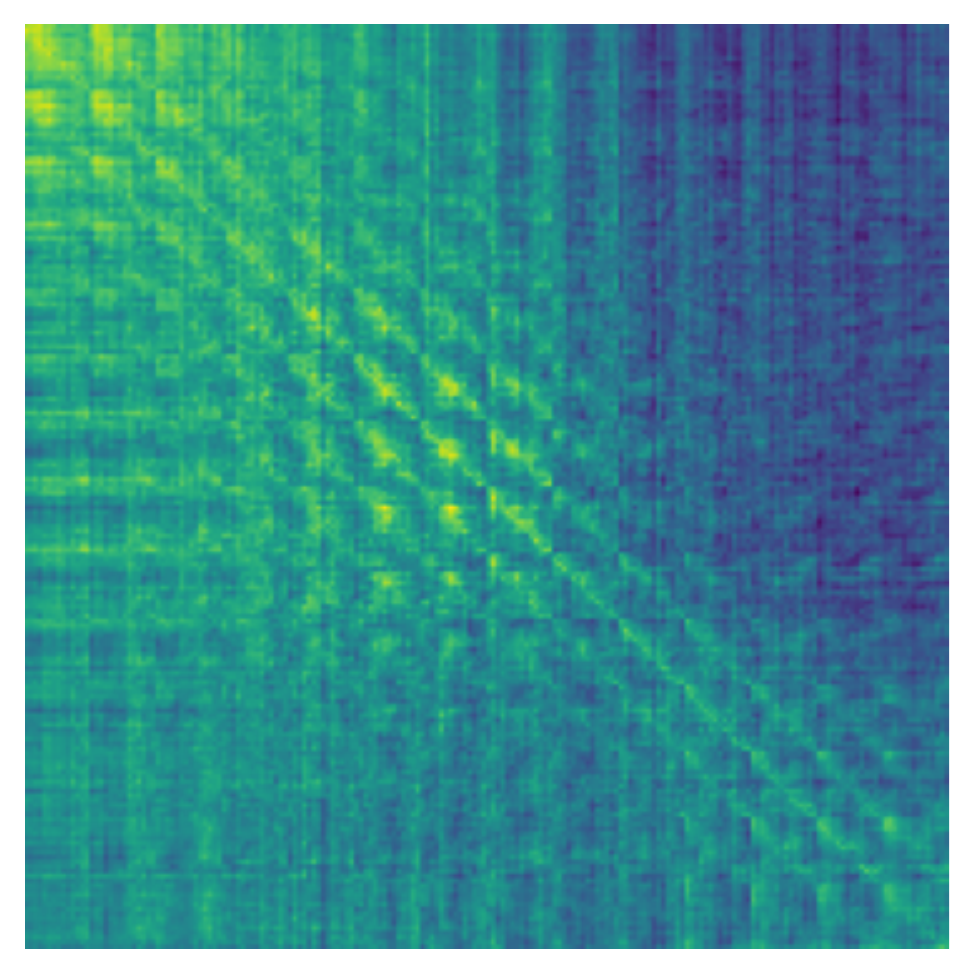}}
	\hfill
	\subfigure{\includegraphics[width=0.15\textwidth,trim=0 0 0 0,clip]{fig/attn/10-cifar-trained.pdf}}
	\hfill
	\subfigure{\includegraphics[width=0.15\textwidth,trim=0 0 0 0,clip]{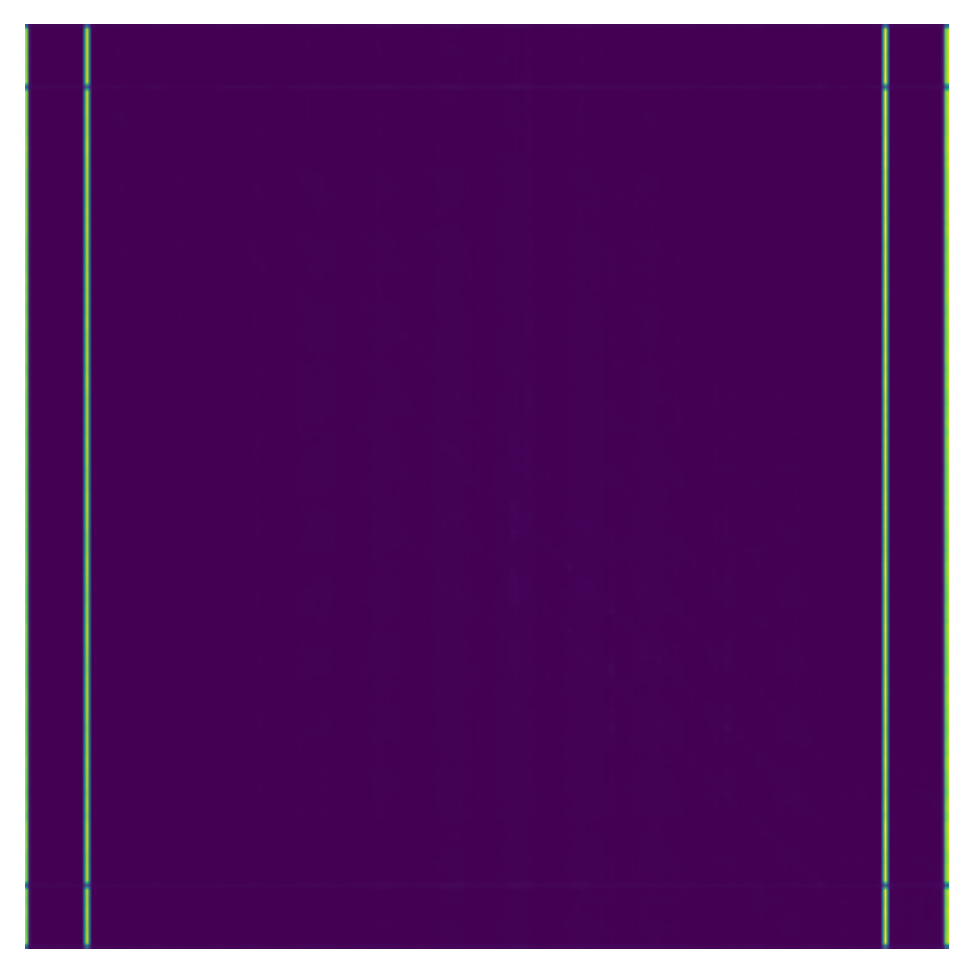}}
	\hfill
	\subfigure{\includegraphics[width=0.15\textwidth,trim=0 0 0 0,clip]{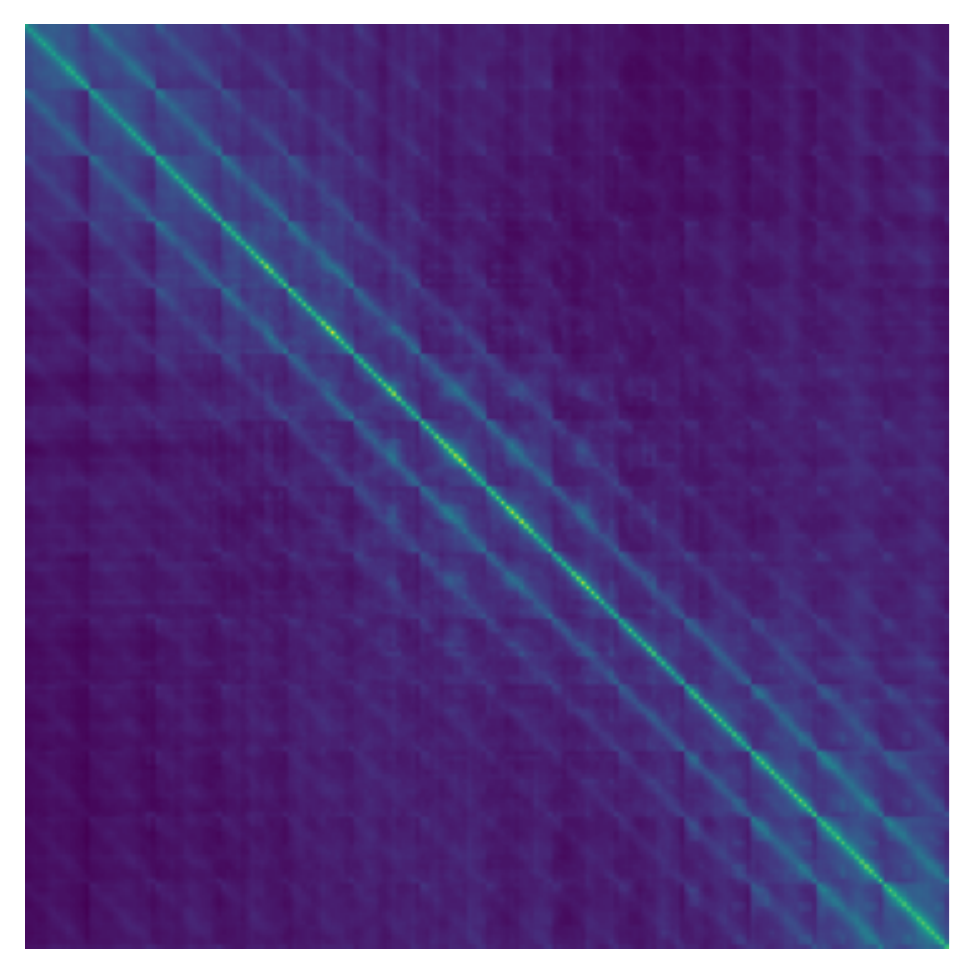}}
	\hfill
	\subfigure{\includegraphics[width=0.15\textwidth,trim=0 0 0 0,clip]{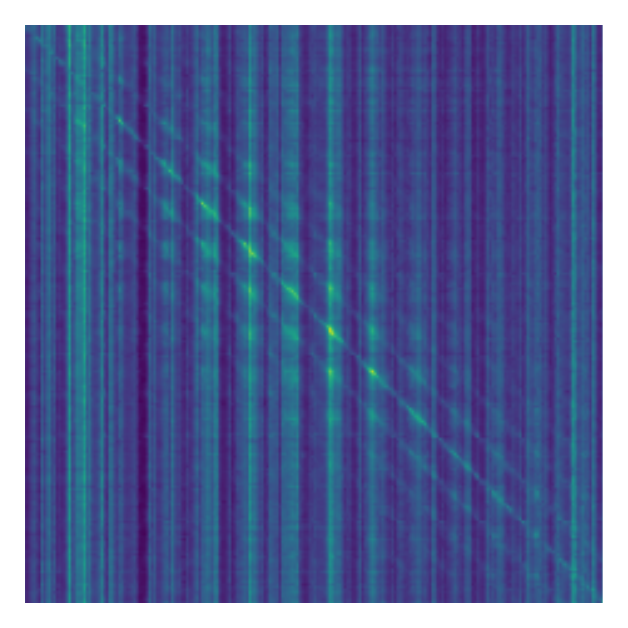}}
    %\vspace*{-0.5em}
    \newline
	\caption{Attention maps computed from one CIFAR-10 batch for ViT-Tiny (plain, pretrained on ImageNet, using our init, and trained on ImageNet using our init). Interestingly, we see that the broadcasting or pooling behavior seen in the uniformly-initialized pretrained model does not occur as clearly in the pretrained model that used our init. \newline Rows: $\downarrow$ Layers \#1, 3, 5, 7, 9, 11}
	\label{ref:attn-maps-apx}
\end{figure}

\clearpage

\begin{figure*}
\centering
\subfigure[$W_Q W_K^T$ does not have such prominent positive diagonals as in an ImageNet-pretrained model. $\rightarrow$ Layers 1-12, $\downarrow$ Attention Heads 1-3 of 12 ]{\includegraphics[width=\textwidth,trim=0 0 0 0,clip]{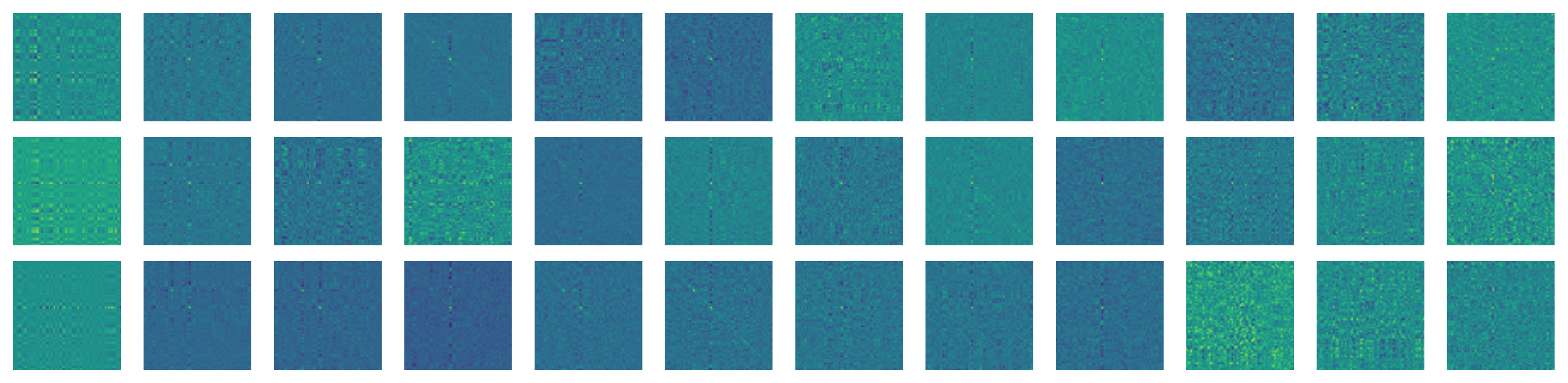}}
    \subfigure[$W_V W_{proj}$ has more faint negative diagonals than in a pretrained model.]{\includegraphics[width=\textwidth,trim=0 0 0 0,clip]{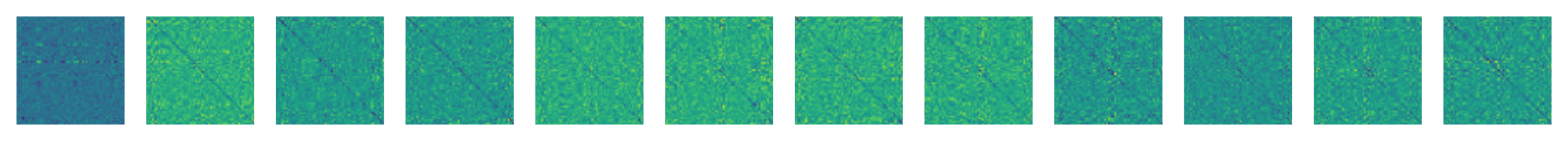}}
\vspace*{-0.5em}
\caption{Training a ViT-Tiny from scratch without our initialization on CIFAR-10 does not show such prominent diagonals in weight products. }
\label{ref:tmr-observations}
\end{figure*}

\begin{figure}
\centering
    \subfigure[Tuning $W_V W_{proj}$]{\includegraphics[height=3.5cm,trim=5 0 55 5,clip]{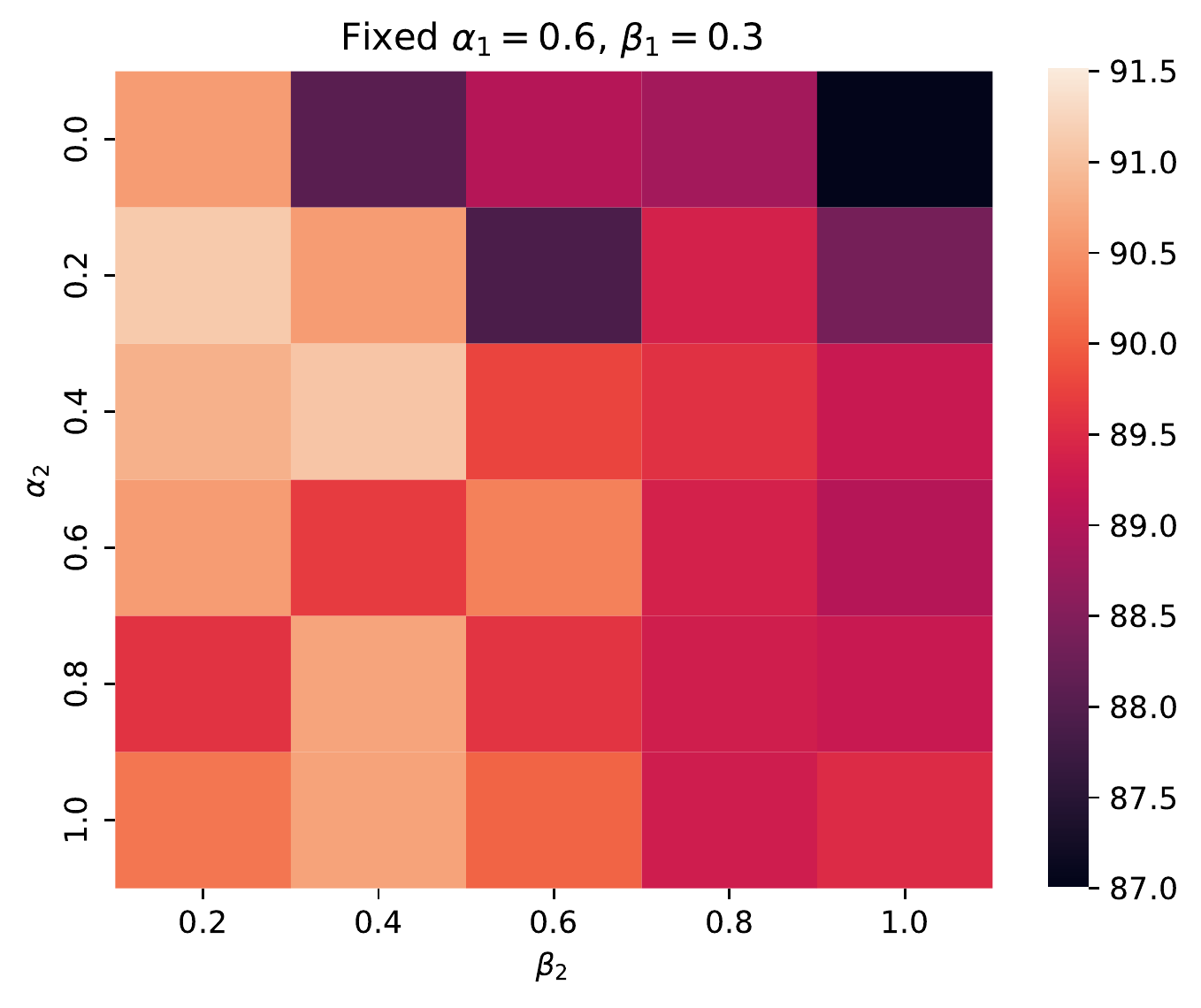}}
    \subfigure[Tuning $W_Q W_K^T$]{\includegraphics[height=3.5cm,trim=5 0 0 5,clip]{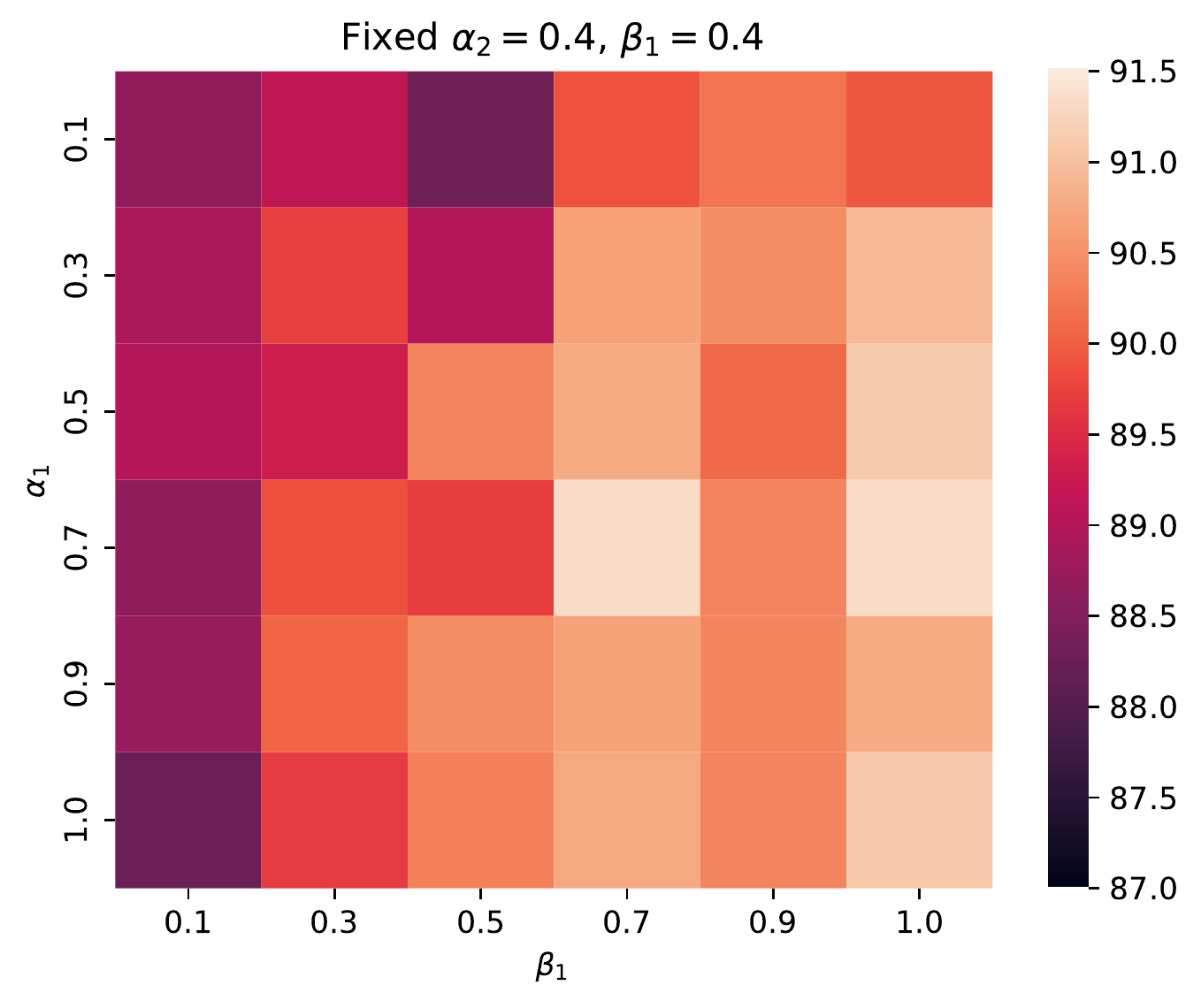}}
\vspace*{-0.5em}
    \caption{Grid search of $\alpha, \beta$ for both $W_V W_{proj}$ and $W_Q W_K^T$ on CIFAR-10, 100 epochs on DeiT-Ti.}
	\label{fig:tuning}
\end{figure}

\clearpage

\begin{table}
\caption{Different internal resolutions}
\label{table:internal-res}
\vskip 0.15in
\begin{center}
\begin{small}
\begin{tabular}{lcccr}
\toprule
\thead{Patch \\ Size} & \thead{Input \\ Size} & \thead{Acc. \\ (Base)} & \thead{Acc. \\ (Init)} & \thead{$\Delta$ Acc.} \\
\midrule
2 & 32 & 85.79 & 90.46 & 4.67 \\
4 & 64 & 90.24 & 92.38 & 2.14 \\
8 & 128 & 90.03 & 92.49 & 2.46 \\
16 & 256 & 88.85 & 92.74 & 3.89 \\
\midrule
4 & 32 & 88.43 & 90.47 & 2.03 \\
8 & 64 & 88.00 & 90.96 & 2.96 \\
16 & 128 & 87.90 & 91.90 & 4.00 \\
32 & 256 & 86.27 & 90.15 & 3.88 \\
\bottomrule
\end{tabular}
\end{small}
\end{center}
\vskip -0.1in
\end{table}

%
%You can have as much text here as you want. The main body must be at most $8$ pages long.
%For the final version, one more page can be added.
%If you want, you can use an appendix like this one, even using the one-column format.
%%%%%%%%%%%%%%%%%%%%%%%%%%%%%%%%%%%%%%%%%%%%%%%%%%%%%%%%%%%%%%%%%%%%%%%%%%%%%%%
%%%%%%%%%%%%%%%%%%%%%%%%%%%%%%%%%%%%%%%%%%%%%%%%%%%%%%%%%%%%%%%%%%%%%%%%%%%%%%%

\end{document}